\documentclass[twocolumn,letterpaper]{article}

\usepackage{cvpr}              

\usepackage{colortbl}
\usepackage{xcolor}
\usepackage[accsupp]{axessibility} 
\definecolor{cvprblue}{rgb}{0.21,0.49,0.74}
\usepackage[pagebackref,breaklinks,colorlinks,allcolors=cvprblue]{hyperref}

\def\paperID{****} 
\def\confName{CVPR}
\def\confYear{2026}

\title{Hierarchical Pre-Training of Vision Encoders with Large Language Models}

\author{
    Eugene Lee$^{1}$, Ting-Yu Chang$^{2}$, Jui-Huang Tsai$^{2}$, Jiajie Diao$^{1}$, Chen-Yi Lee$^{2}$ \\
    $^{1}$ University of Cincinnati \quad
    $^{2}$ National Yang Ming Chiao Tung University \\
    {\tt\small eugene.lee@uc.edu, jiajie.diao@uc.edu, cylee@nycu.edu.tw}
    }
\newcommand\blfootnote[1]{%
  \begingroup
  \renewcommand\thefootnote{}\footnote{#1}%
  \addtocounter{footnote}{-1}%
  \endgroup
}
\begin{document}
\maketitle

\begin{abstract}
The field of computer vision has experienced significant advancements through scalable vision encoders and multimodal pre-training frameworks. However, existing approaches often treat vision encoders and large language models (LLMs) as independent modules, limiting the integration of hierarchical visual features. In this work, we propose HIVE (Hierarchical Pre-Training of Vision Encoders), a novel framework that enhances vision-language alignment by introducing hierarchical cross-attention between the vision encoder and LLM. Unlike conventional methods that flatten image embeddings, HIVE enables structured feature fusion across multiple layers, improving gradient flow and representation learning. To optimize this interaction, we introduce a three-stage training strategy that progressively aligns the vision encoder with the LLM, ensuring stable optimization and effective multimodal fusion. Empirical evaluations demonstrate that HIVE achieves superior performance not only in image classification but also on various vision-language tasks, outperforming self-attention-based methods in benchmarks such as MME, GQA, OK-VQA, and ScienceQA. Our results highlight the benefits of hierarchical feature integration, paving the way for more efficient and expressive vision-language models.
\end{abstract}
    
\begin{figure}[t]
    \centering
    \includegraphics[width=\linewidth]{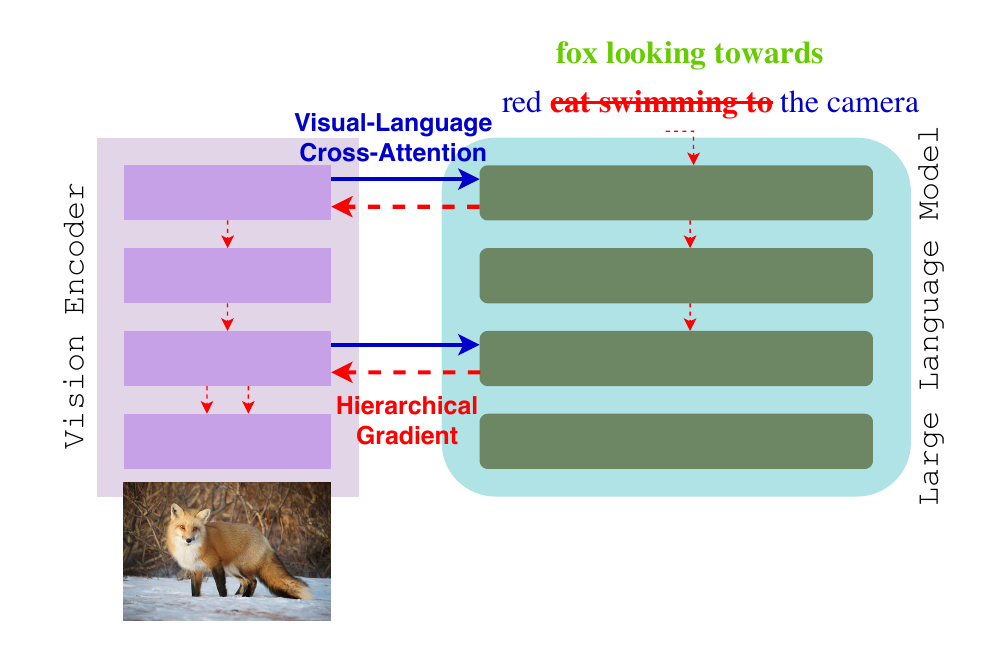}
    \caption{
        Overview of the proposed Hierarchical Pre-Training of Vision Encoders (HIVE) framework. The vision encoder extracts hierarchical visual features, which are projected and integrated into the large language model (LLM) through cross-attention. This enables multi-layered feature alignment, improving representation learning, semantic understanding, and gradient flow. Green text indicates correct model predictions, while red text highlights incorrect predictions.
    }
    \label{fig:hive-overview}
\end{figure}

\section{Introduction}  
\blfootnote{Project page and code: \url{https://eugenelet.github.io/HIVE-Project/}}
\label{sec:intro}  

The field of computer vision has advanced rapidly through scalable vision encoders and multimodal pre-training. Models like AIMv2~\cite{fini2024multimodal} leverage autoregressive pre-training to achieve strong performance across diverse benchmarks. However, a critical challenge remains: existing approaches typically feed flattened image embeddings directly into the input layer of large language models (LLMs). This shallow integration restricts the utilization of hierarchical visual features and limits gradient flow across model layers.  

While LLMs exhibit exceptional natural language capabilities~\cite{zhou2022learning, nie2023pro}, and multimodal models like CLIP effectively align vision and language, they often treat vision encoders and LLMs as largely independent modules. Relying purely on the LLM's self-attention to process visual inputs fails to fully exploit the rich, multi-level features extracted by the vision encoder, thereby reducing representational power and downstream performance.

Alternative strategies have sought to bridge this gap. BLIP-2~\cite{li2023blip} introduced a Q-Former to extract a small set of visual features for a frozen LLM, reducing computational overhead but restricting direct cross-modal interaction. Similarly, Flamingo~\cite{alayrac2022flamingo} proposed gated cross-attention to incorporate visual information while maintaining frozen language capabilities. However, these methods primarily optimize the LLM's visual processing rather than explicitly improving the vision encoder’s ability to learn hierarchical representations.

Beyond CLIP, diverse strategies aim to enhance image representations before LLM integration. Ramesh et al.~\cite{ramesh2022hierarchical} utilized CLIP latents for image generation, while Yu et al.~\cite{yu2022coca} leveraged contrastive captioners. He et al.~\cite{he2022masked} demonstrated the efficacy of masked autoencoders (MAEs) for scalable vision learning, emphasizing hierarchical feature extraction. Recent works have also explored autoregressive multimodal modeling~\cite{lu2024unified} and self-supervised visual representation learning like SimCLR~\cite{chen2020simple}. Despite these advancements, standard approaches still underutilize the vision encoder's hierarchical structure during LLM interaction. 

To address these limitations, we propose Hierarchical Pre-Training of Vision Encoders (HIVE). Instead of feeding flattened image embeddings into the LLM, HIVE establishes multi-layered cross-attention between the vision encoder and the LLM. This enables dynamic interaction between hierarchical visual features and language representations, preserving fine-grained spatial information, enhancing gradient flow, and improving downstream generalization. Figure~\ref{fig:hive-overview} illustrates our framework.

To stabilize this hierarchical integration, we refine a \textit{three-stage training strategy} inspired by prior multimodal pipelines~\cite{lu2024deepseek, li2024llava2}. Stage 1 trains a lightweight projector to map visual representations into the LLM’s input space while keeping both base models frozen. Stage 2 jointly optimizes the projector and LLM to adapt to multi-level visual features. Finally, Stage 3 fine-tunes all components end-to-end, allowing the vision encoder to align seamlessly with the hierarchical cross-attention mechanism. This progressive optimization successfully mitigates early-stage misalignment.

Our contributions are summarized as follows:  
\begin{enumerate}  
    \item \textit{Hierarchical Cross-Attention}: We propose a novel framework that integrates multi-level vision encoder features into the LLM, improving visual-text alignment and feature propagation.  
    \item \textit{Adapted Three-Stage Training}: We refine an established training pipeline to stabilize hierarchical cross-attention, ensuring efficient, structured vision-language interactions.  
    \item \textit{Enhanced Performance and Efficiency}: HIVE consistently outperforms self-attention baselines across classification and vision-language benchmarks while significantly reducing computational overhead.
\end{enumerate}  

Extensive experiments showcase HIVE's robust performance gains. The remainder of this paper is structured as follows: Section~\ref{sec:related} reviews related work. Section~\ref{sec:method} details our cross-attention framework and training strategy. Section~\ref{sec:experiments} presents experimental results, and Section~\ref{sec:conclusion} concludes the paper.

\section{Related Work}  
\label{sec:related}  

\paragraph{Vision Encoders and Hierarchical Feature Integration}  
Vision encoders such as CLIP~\cite{radford2021learning} and SigLIP~\cite{zhai2023sigmoid} have shown the potential of aligning visual and textual representations through contrastive pre-training, achieving state-of-the-art results across various benchmarks. AIMv2~\cite{fini2024multimodal}, a leading vision encoder, introduced autoregressive pre-training that effectively processes image patches and text tokens, enabling robust performance on both vision and multimodal tasks. However, a key limitation of these approaches lies in their reliance on flattened image embeddings as inputs to large language models (LLMs), which neglects the hierarchical nature of features extracted by vision encoders. Recent efforts, such as Rezaei et al.~\cite{rezaei2024learning}, leveraged self-supervised learning to improve attention mechanisms within transformer-based vision encoders, while Pan et al.~\cite{pan2025tokenize} proposed tokenization methods to enable fine-grained feature extraction.  

Beyond standard pre-training approaches, models such as Alpha-CLIP~\cite{sun2024alpha} and LexLIP~\cite{luo2023lexlip} have explored targeted enhancements to CLIP-based encoders by refining local feature extraction and lexicon-bottlenecked representations, respectively. Similarly, studies such as Baldrati et al.~\cite{baldrati2022effective} and Sain et al.~\cite{sain2023clip} have investigated novel retrieval-based mechanisms that enhance compositional image-text alignment. These models illustrate the importance of structured vision encoders, but they still lack explicit hierarchical feature utilization, which remains underexplored, particularly in pre-training frameworks where cross-modal interactions play a significant role in improving representational capacity.

In this context, Alabdulmohsin et al.~\cite{alabdulmohsin2023getting} proposed a scalable ViT framework that optimizes model design for computational efficiency while maintaining robust performance. Additionally, Barbu et al.~\cite{barbu2019objectnet} introduced ObjectNet, a dataset that systematically evaluates object recognition models by controlling for bias, providing valuable insights for improving generalization in vision models.

\paragraph{Vision-Language Models and Cross-Attention Mechanisms}  
Vision-language models extend the capabilities of vision encoders by introducing mechanisms for cross-modal alignment and reasoning. For instance, LLaVA-o1~\cite{xu2024llava} and Qwen2-VL~\cite{wang2024qwen2} improved perception and reasoning capabilities by incorporating high-resolution visual processing and step-by-step reasoning strategies. Generative multimodal models such as Sun et al.~\cite{sun2024generative} have demonstrated in-context learning capabilities that enhance multimodal reasoning, while Bai et al.~\cite{bai2023qwen} leveraged instruction-tuning frameworks to handle complex multimodal tasks, including text reading and object localization.  

In addition to these developments, Qwen-VL~\cite{bai2023qwen} explored versatile vision-language capabilities through improved visual encoding for multimodal understanding. Similarly, studies like Alabdulmohsin et al.~\cite{alabdulmohsin2024clip} have addressed the challenges of balancing data during multimodal learning, while Evans et al.~\cite{evans2024data} introduced data curation methods that accelerate multimodal learning via example selection strategies.  

While these approaches have achieved notable success, they often treat vision encoders and LLMs as independent components, relying on simplistic integration strategies that fail to exploit the rich, hierarchical features generated by vision encoders. Cross-attention mechanisms, which enable deep feature interaction across multiple layers of vision encoders and LLMs, have shown promise in improving such integrations. For example, techniques such as TCP~\cite{yao2024tcp} and CoOp~\cite{zhou2022learning} have demonstrated that task-specific prompts can improve alignment between modalities. However, these methods lack the depth to incorporate multi-layer hierarchical representations, highlighting the need for more robust pre-training frameworks that utilize cross-attention to enhance gradient flow and semantic alignment.  

Several recent approaches have focused on leveraging frozen vision encoders while training LLMs for vision-language tasks. BLIP-2~\cite{li2023blip} introduced a querying transformer (Q-Former) that extracts a small number of informative visual features before passing them to a frozen LLM, effectively creating a vision-language interface with minimal computational overhead. Similarly, Flamingo~\cite{alayrac2022flamingo} introduced gated cross-attention layers between a frozen vision encoder and a pre-trained LLM, allowing vision-language alignment through interleaved image-text sequences. Methods such as UMG-CLIP~\cite{shi2024umg} focus on multi-granularity processing for open-world vision tasks, whereas models like Tong et al.~\cite{tong2024eyes} have critically analyzed the limitations of multimodal vision-language models in terms of their visual understanding capabilities.  

While these methods demonstrate strong zero-shot and few-shot learning capabilities, their primary objective is LLM training, rather than vision encoder pre-training. In contrast, the proposed HIVE framework explicitly pre-trains the vision encoder, ensuring that hierarchical features are effectively captured before interacting with the LLM. This distinction allows HIVE to fully utilize multi-layered representations, improving both feature extraction and downstream classification performance.

\paragraph{Scalability and Challenges in Multimodal Pre-Training}  
Scalability and computational efficiency remain critical concerns in the development of multimodal pre-training frameworks. AIMv2~\cite{fini2024multimodal} demonstrated the scalability of autoregressive pre-training for large datasets, while models like Qwen2-VL~\cite{wang2024qwen2} explored efficient strategies for processing high-resolution images without compromising performance. Meanwhile, prompt-based approaches such as Pro-tuning~\cite{nie2023pro} and Kan et al.~\cite{kan2023knowledge} introduced mechanisms to enhance task-specific alignment through textual prompts, improving adaptability for diverse applications.  

Models such as FlexiViT~\cite{beyer2023flexivit} have demonstrated compute-optimal designs for scalable vision transformers, while Dehghani et al.~\cite{dehghani2023patch} explored flexible visual transformer architectures that efficiently process arbitrary input resolutions. Similarly, Fan et al.~\cite{fan2023improving} presented effective language rewrite mechanisms to enhance CLIP training for improved vision-language alignment.  

Furthermore, Beyer et al.~\cite{beyer2020we} explored the limitations of ImageNet-scale datasets and proposed enhancements for scalable model designs, while PaliGemma~\cite{beyer2024paligemma} introduced a versatile vision-language model for improved multimodal transfer capabilities.  

Nevertheless, these methods often neglect the computational complexities associated with integrating hierarchical visual features across vision encoders and LLMs. Pan et al.~\cite{pan2025tokenize} addressed part of this challenge through tokenization strategies that preserve fine-grained features, but this approach lacks the scalability required for large-scale cross-modal training. Wang et al.~\cite{wang2024diffusion} proposed DIVA, which refines CLIP’s visual representations using generative feedback from diffusion models. However, DIVA operates purely in the vision domain without leveraging vision-language datasets or cross-attention between vision encoders and LLMs.  

Several generative and retrieval-based models have attempted to enhance cross-modal scalability, such as CLIPDraw~\cite{frans2022clipdraw} and Text2LIVE~\cite{bar2022text2live}, which focus on leveraging multimodal encoders for image and video manipulation. Crowson et al.~\cite{crowson2022vqgan} explored vision-language editing mechanisms using VQGAN-CLIP, further illustrating how CLIP-derived architectures can be applied across various creative domains.  

In contrast, the proposed method integrates cross-attention between vision and language modalities, effectively capturing hierarchical features while benefiting from large-scale vision-language datasets. This approach enables richer multimodal understanding and better scalability in vision-language tasks.

\section{Method}  
\label{sec:method}  

In this section, we introduce the proposed hierarchical cross-attention framework for pre-training vision encoders with large language models (LLMs). The core idea is to establish hierarchical interactions between the vision encoder and LLM by integrating multi-level visual features into the LLM through cross-attention. Unlike traditional approaches that flatten image features into a single vector input, our method enables the LLM to process structured visual representations across multiple levels of abstraction. This results in improved cross-modal alignment and enhances the vision encoder's representational capacity for downstream tasks.  

\subsection{Overview}  
The proposed framework employs a hierarchical cross-attention mechanism to progressively align vision encoder features with the LLM. Instead of directly feeding a single-level feature representation, multiple layers from the vision encoder are projected into the LLM, allowing it to attend to both low-level details and high-level semantic concepts. This enables fine-grained feature integration that preserves spatial and structural information.  

Given an input image $\mathbf{I} \in \mathbb{R}^{H \times W \times C}$, the vision encoder extracts hierarchical features at different depths, which are subsequently mapped via a lightweight projector before being attended to by the LLM. This hierarchical integration allows the LLM to process visual information at multiple levels, enhancing its capability to reason about fine-grained object structures and high-level scene semantics. The overall architecture is shown in Figure~\ref{fig:hive-overview}.  

\subsection{Hierarchical Feature Integration}  
\label{subsec:hierarchical_integration}  

The input image $\mathbf{I}$ is tokenized into a sequence of visual tokens $\mathbf{T}_v = [\mathbf{t}_1, \mathbf{t}_2, \ldots, \mathbf{t}_N]$, where $N$ is the number of patches. The vision encoder processes these tokens through multiple layers, generating a sequence of hierarchical feature representations:  
$$\mathbf{F}_l = f_l(\mathbf{F}_{l-1}), \quad l = 1, \ldots, L$$
where $\mathbf{F}_0 = \mathbf{T}_v$ and $f_l$ represents the transformation function at encoder layer $l$. To prevent dimension mismatch and parameter explosion, we do not project every layer. Instead, we define a selection subset $\mathcal{S} \subset \{1, \dots, L\}$ representing a 25\% uniform sampling of the encoder depths. 

A lightweight projector function $g_l$ maps these selected hierarchical features to a dimension compatible with the LLM:
$$\mathbf{T}_{\text{LLM}, l} = g_l(\mathbf{F}_l), \quad l \in \mathcal{S}$$
The function $g_l$ is implemented as a 2-layer multi-layer perceptron (MLP) with a GELU activation and residual connections. This ensures that projected features retain their structural integrity without introducing severe training bottlenecks.

\subsection{Hierarchical Cross-Attention}  
\label{subsec:cross_attention}  

To enable structured feature fusion, the projected vision features interact with the LLM through a hierarchical cross-attention mechanism. Specifically, we inject cross-attention layers into the LLM at depths corresponding to our selected subset $\mathcal{S}$. 

For a given aligned layer $l$, the interaction is formulated as:  
$$\mathbf{H}_l = \text{CrossAttention}(\mathbf{Q}_l, \mathbf{K}_l, \mathbf{V}_l)$$
where the query $\mathbf{Q}_l$ is derived strictly from the LLM's intermediate hidden states at layer $l$. The key $\mathbf{K}_l$ and value $\mathbf{V}_l$ matrices are linearly projected from the corresponding vision encoder features $\mathbf{T}_{\text{LLM}, l}$. The attention weights are computed as:
$$\mathbf{A}_l = \text{Softmax}\left(\frac{\mathbf{Q}_l \mathbf{K}_l^T}{\sqrt{d}}\right)$$
where $d$ is the feature dimension. By physically routing intermediate vision features to intermediate LLM layers, we force the vision encoder to preserve low-level structural data (like edges and textures) in its early layers, and abstract semantics in its deeper layers, directly supervised by the LLM's language modeling loss.

\subsection{Training Optimization}  
The model is trained using the next-token prediction cross-entropy loss:
\begin{equation}
    \mathcal{L} = -\sum_{t} p_t \log \hat{p}_t,
\end{equation}
where $p_t$ represents the ground truth token probability, and $\hat{p}_t$ is the predicted token distribution from the LLM.  

The three-stage training procedure progressively optimizes different components of the model:
- Stage 1: The function $g_l$ (projector) is trained while keeping the vision encoder and LLM frozen, ensuring that vision features are mapped correctly into the LLM's token space.  
- Stage 2: The function $g_l$ and the LLM are jointly trained while keeping the vision encoder frozen, allowing the LLM to refine its feature processing.  
- Stage 3: The entire model, including $f_l$ (vision encoder), is trained end-to-end to fully optimize hierarchical feature interactions.  

This staged approach ensures stable training dynamics and prevents early-stage misalignment between the vision encoder and LLM. The hierarchical cross-attention mechanism enables the LLM to process rich, structured visual representations, leading to improved cross-modal understanding.

\subsection{Three-Stage Training Procedure}  
\label{subsec:training_procedure}

The training process follows a structured three-stage progression to align the vision encoder $f_l$, projector $g_l$, and large language model (LLM) $L$ while ensuring stable optimization and effective cross-modal interactions. This staged approach has been widely adopted in vision-language training pipelines, as demonstrated in DeepSeek-VL~\cite{lu2024deepseek} and LLaVA~\cite{li2024llava,li2024llava2}. Inspired by these works, we implement a tailored variant of this procedure for hierarchical cross-attention-based multimodal learning. Because the massive parameter space of hierarchical cross-attention is prone to catastrophic forgetting and gradient explosion if optimized from scratch, we do not ablate the removal of the projection warm-up stages. Extensive prior work~\cite{lu2024deepseek, li2024llava} has empirically established that unaligned vision-language joints fail to converge without this three-stage progressive unlocking.

\paragraph{Stage 1: Projector Pre-Training}  
In the first stage, only the projector $g_l$ is trained, while both the vision encoder $f_l$ and the LLM $L$ remain frozen. The vision encoder extracts hierarchical features $\mathbf{F}_l$, which are mapped by $g_l$ into the LLM’s token space as $\mathbf{T}_{\text{LLM}, l}$. This step ensures that projected representations align with the LLM’s embedding space before deeper integration. The LLM processes these features without updates, with optimization driven by the next-token prediction loss.

\paragraph{Stage 2: Joint Training of LLM and Projector}  
With the vision encoder still frozen, the second stage optimizes both the projector $g_l$ and the LLM $L$. This step refines the LLM’s ability to utilize projected hierarchical features while allowing the projector to better adapt to the LLM’s internal representations. The projected representations $\mathbf{T}_{\text{LLM}, l}$ serve as inputs to the LLM, which generates token-level outputs $\mathbf{O}$. This stage facilitates the development of stronger cross-modal alignment, as similarly explored in prior vision-language models~\cite{li2024llava}.

\paragraph{Stage 3: Joint Training of All Components}  
In the final stage, all components—vision encoder $f_l$, projector $g_l$, and LLM $L$—are trained together to fully optimize hierarchical cross-modal interactions. The vision encoder is now updated to generate features that are maximally informative for LLM integration. Cross-attention is applied across multiple layers of both models, enabling structured information exchange between vision and language modalities. The entire pipeline is optimized end-to-end using the same next-token prediction loss, ensuring cohesive representation learning across modalities.

\subsection{Fine-Tuning and Evaluation}
After completing the three-stage pre-training process, the model is fine-tuned for both classification and vision-language model (VLM) tasks to ensure optimal performance.

For classification tasks, a lightweight classifier head is added to the end of the frozen vision encoder. During this stage, only the classifier head is trained, while the vision encoder itself remains frozen. Since the projector and LLM are used solely during the vision encoder's pre-training phase, they are not involved in the classification fine-tuning process. This streamlined approach efficiently adapts the vision encoder’s learned representations to downstream classification tasks.

For vision-language tasks, we adopt a two-stage fine-tuning strategy inspired by LLaVA~\cite{li2024llava,li2024llava2}. In the first stage, only the connector (projector) is trained while keeping the vision encoder and LLM frozen. In the second stage, the vision encoder remains frozen, and only the LLM is updated. This staged fine-tuning procedure enhances the LLM’s ability to leverage visual features for downstream multimodal tasks while maintaining stable and efficient optimization.

\paragraph{Rationale for Downstream Architecture Shift:} A natural question arises as to why hierarchical cross-attention is employed during pre-training, but standard token concatenation (LLaVA-style) or linear probing is used during downstream fine-tuning. This is a deliberate design choice optimized for inference efficiency. Our primary hypothesis is that hierarchical cross-attention acts as a rigorous \textit{pre-training regularization objective}. By forcing the LLM to attend to intermediate vision layers during the pre-training phase, we explicitly supervise the vision encoder to retain dense, multi-scale semantic information. Once the vision encoder is fully trained, its final-layer representations are inherently richer and more structurally sound. Therefore, during downstream fine-tuning, this enriched encoder can simply feed its final-layer tokens into a standard architecture. This transfers the representational benefits of hierarchical pre-training without incurring the heavy computational overhead of multi-layer cross-attention during deployment and inference.

\subsection{Computational Complexity Analysis}
\label{subsec:complexity_summary}

We summarize the computational complexity of our proposed hierarchical cross-attention framework compared to traditional self-attention mechanisms in vision-language models (VLMs).

In conventional self-attention, both the attention mechanism and the MLP layers operate across the entire set of visual and text tokens, resulting in a complexity of:
\begin{equation}
    \mathcal{O} \left( L_l \frac{N^2 d}{2} + L_l N d^2 \right)
\end{equation}
where $N = N_v + N_t$ is the total number of vision and text tokens. The quadratic term $N^2$ arises from the pairwise token interactions in self-attention, while the MLP complexity term scales linearly with $N$.

In contrast, our proposed hierarchical cross-attention mechanism selectively integrates multi-level visual features into the LLM. By bypassing the LLM’s MLP layers for visual tokens and only passing a subset of features to the LLM, the complexity is reduced to:
\begin{equation}
    \mathcal{O} (L_l L_s d^2 + L_l N_t d^2)
\end{equation}
where $L_s \ll N_v$ is the number of selected vision encoder layers. This design significantly reduces computational overhead by eliminating redundant MLP computations and limiting cross-modal interactions to a smaller feature set.

By reducing both the number of visual tokens processed and the overhead introduced by MLP layers, hierarchical cross-attention achieves significant efficiency improvements over full self-attention. These computational savings enable our framework to scale effectively while maintaining strong performance across diverse visual and multimodal tasks. For a detailed derivation and complexity comparison, please refer to the appendix.
\section{Experiments} \label{sec:experiments}

We evaluate our hierarchical cross-attention framework (HIVE) through comprehensive experiments. Our evaluation covers benchmark comparisons against self-attention baselines, gradient flow analysis to study optimization dynamics, and attention map visualization to highlight improved feature alignment. We also conduct an efficiency analysis demonstrating the computational benefits of HIVE. Detailed results, including complexity analysis, are provided in the Appendix.

\subsection{Experimental Setup}
\label{subsec:setup}

\paragraph{Datasets.}  
For classification tasks, we evaluate our method on CIFAR-10, CIFAR-100, ImageNet-1K, Tiny-ImageNet, Food-101, Stanford Cars, Oxford-IIIT Pets, and Caltech-256. These datasets cover a diverse range of scales and complexities, ensuring robust performance assessment.

For vision-language model (VLM) evaluation, we use MME, GQA, OK-VQA, and ScienceQA, which assess visual reasoning, object understanding, and multimodal knowledge grounding.

\paragraph{Baselines.}  
We compare HIVE against two baseline configurations: 
\begin{itemize}
    \item \textit{Base}: The original foundation models CLIP (clip-vit-large-patch14-336) and SigLIP (siglip-large-patch16-384) without additional LLM-supported pre-training.
    \item \textit{SA}: A self-attention-based vision encoder trained using our three-stage pre-training method.
\end{itemize}

Both SA and HIVE follow identical pre-training strategies for the vision encoder to ensure a fair comparison. For vision-language model (VLM) evaluation, both SA and HIVE are further fine-tuned following the procedure used in LLaVA~\cite{liu2024visual}. These baselines provide a strong benchmark to assess the benefits of hierarchical cross-attention over conventional self-attention-based models.

\paragraph{Implementation Details.}  
For pre-training, we use MobileLLM-350M \cite{liu2024mobilellm} as the language model backbone for both self-attention and hierarchical cross-attention configurations. Optimization is performed using decoupled AdamW \cite{loshchilov2017decoupled} with a peak learning rate of $1\times10^{-3}$ and a cosine decay schedule. To maintain stability, we apply gradient clipping and a linear warmup phase. For VLM fine-tuning, we adopt LLaVA using the Llama-3.2-1B-Instruct model following standard VLM training practices and hyperparameters.

For classification tasks, we append a classifier head to the vision encoder and fine-tune only the classifier while keeping the vision encoder frozen.

All experiments are conducted on a single RTX 3090 GPU due to limited computational resources. Due to this hardware constraint (24GB VRAM), we utilize a maximum batch size of 256 for the early stages and accumulate gradients where necessary. This computational context also informs our targeted approach to ablations, prioritizing theoretically sound training stages over exhaustive grid searches.

\subsection{Comparison with Self-Attention Methods}  
\label{subsec:comparison}

\paragraph{Vision Encoder Classification.}
We evaluate the classification performance of our proposed hierarchical cross-attention framework, HIVE, against self-attention-based methods. Our primary objective is to assess the effectiveness of the pre-trained vision encoder in extracting high-quality visual representations. As shown in Table~\ref{tab:comparison}, HIVE consistently outperforms self-attention-based baselines across multiple datasets. Notably, the most significant improvements are observed on fine-grained classification benchmarks such as \textbf{Food-101}, \textbf{Caltech-256}, and \textbf{Pets} (SigLIP-based), demonstrating HIVE's ability to capture intricate visual details. Moreover, HIVE's strong performance on \textbf{Tiny-ImageNet} highlights its robust generalization across datasets of varying sizes and complexity. These results validate the effectiveness of our hierarchical cross-attention design in enhancing visual feature extraction for diverse classification tasks.

\begin{table*}[ht]
    \centering
    \small
\caption{Classification accuracy (\%) on various benchmarks comparing our proposed method, HIVE, with self-attention-based approaches. Results are reported as mean ± standard deviation (SD) over 3 runs. \textbf{Base} represents the original foundation models (CLIP or SigLIP) without additional LLM-supported pre-training. \textbf{SA} represents a self-attention-based vision encoder with LLM-supported pre-training, while \textbf{HIVE} represents our proposed cross-attention-based pre-training framework. The best results are in bold.}
    \label{tab:comparison}
    \begin{tabular}{lcccccccc}
        \toprule
        Method & CIFAR-10 & CIFAR-100 & ImageNet-1K & Tiny-ImageNet & Food-101 & Cars & Pets & Caltech-256 \\
        \midrule
        \multicolumn{9}{l}{\textbf{CLIP-based Models}} \\  
        \cmidrule(lr){1-9}
        Base         
        & 98.32\textsuperscript{$\pm$0.04} 
        & 87.92\textsuperscript{$\pm$0.08} 
        & 84.01\textsuperscript{$\pm$0.01} 
        & 86.65\textsuperscript{$\pm$0.11} 
        & 95.75\textsuperscript{$\pm$0.03} 
        & \textbf{91.17}\textsuperscript{$\pm$0.04} 
        & \textbf{96.02}\textsuperscript{$\pm$0.10} 
        & 96.12\textsuperscript{$\pm$0.08} \\
        
        SA  
        & 98.37\textsuperscript{$\pm$0.01} 
        & 87.93\textsuperscript{$\pm$0.06} 
        & \textbf{84.14}\textsuperscript{$\pm$0.02} 
        & 86.51\textsuperscript{$\pm$0.04} 
        & 95.77\textsuperscript{$\pm$0.03} 
        & 91.04\textsuperscript{$\pm$0.13} 
        & 95.98\textsuperscript{$\pm$0.05} 
        & 96.27\textsuperscript{$\pm$0.10} \\
        
        \rowcolor{gray!20} HIVE    
        & \cellcolor{gray!20} \textbf{98.49\textsuperscript{$\pm$0.01}}
        & \cellcolor{gray!20} \textbf{88.49\textsuperscript{$\pm$0.07}}
        & \cellcolor{gray!20} 84.08\textsuperscript{$\pm$0.00}
        & \cellcolor{gray!20} \textbf{86.71\textsuperscript{$\pm$0.08}}
        & \cellcolor{gray!20} \textbf{95.78\textsuperscript{$\pm$0.03}}
        & \cellcolor{gray!20} 91.15\textsuperscript{$\pm$0.23}
        & \cellcolor{gray!20} 95.92\textsuperscript{$\pm$0.07}
        & \cellcolor{gray!20} \textbf{96.35\textsuperscript{$\pm$0.11}} \\
        
        \midrule
        \multicolumn{9}{l}{\textbf{SigLIP-based Models}} \\  
        \cmidrule(lr){1-9}
        Base          
        & 98.43\textsuperscript{$\pm$0.01} 
        & 90.08\textsuperscript{$\pm$0.11} 
        & 85.99\textsuperscript{$\pm$0.02} 
        & 82.27\textsuperscript{$\pm$0.20} 
        & 96.55\textsuperscript{$\pm$0.01} 
        & 94.86\textsuperscript{$\pm$0.15} 
        & 96.76\textsuperscript{$\pm$0.04} 
        & 97.27\textsuperscript{$\pm$0.09} \\
        
        SA  
        & 98.42\textsuperscript{$\pm$0.04} 
        & 89.97\textsuperscript{$\pm$0.10} 
        & 86.04\textsuperscript{$\pm$0.07} 
        & 82.46\textsuperscript{$\pm$0.10} 
        & 96.55\textsuperscript{$\pm$0.03} 
        & 94.85\textsuperscript{$\pm$0.20} 
        & 96.76\textsuperscript{$\pm$0.08} 
        & 97.27\textsuperscript{$\pm$0.05} \\
        
        \rowcolor{gray!20} HIVE  
        & \cellcolor{gray!20} \textbf{98.45\textsuperscript{$\pm$0.04}}
        & \cellcolor{gray!20} \textbf{90.19\textsuperscript{$\pm$0.15}}
        & \cellcolor{gray!20} \textbf{86.06\textsuperscript{$\pm$0.02}}
        & \cellcolor{gray!20} \textbf{82.48\textsuperscript{$\pm$0.05}}
        & \cellcolor{gray!20} \textbf{96.56\textsuperscript{$\pm$0.03}}
        & \cellcolor{gray!20} \textbf{95.09\textsuperscript{$\pm$0.19}}
        & \cellcolor{gray!20} \textbf{96.78\textsuperscript{$\pm$0.02}}
        & \cellcolor{gray!20} \textbf{97.33\textsuperscript{$\pm$0.11}} \\
        \bottomrule
    \end{tabular}
\end{table*}

\paragraph{Vision-Language Model Evaluation.}  
To further evaluate the benefits of hierarchical pre-training, we assess the performance of our vision encoder when integrated with a large language model (LLM) on vision-language tasks. Due to the consistent performance gains observed with SigLIP, we focus solely on SigLIP-based models for these experiments to reduce computational overhead.

Both our self-attention (SA) and hierarchical cross-attention (HIVE) models follow our proposed three-stage pre-training method for vision encoder training. For the final vision-language model (VLM) training stage, we adopt the training procedure outlined in LLaVA~\cite{xu2024llava}, where visual features extracted from the pre-trained encoder are fed as token embeddings into the LLM.

As shown in Table~\ref{tab:llm-eval}, pre-training the vision encoder using our hierarchical cross-attention framework leads to significant performance improvements in vision-language tasks. Note that the scores reported for the MME benchmark represent the combined sum of the perception and reasoning sub-tasks. HIVE consistently outperforms both the SigLIP baseline and the SA model across all evaluated benchmarks. The most substantial improvements are observed in Visual Question Answering tasks (GQA and OK-VQA) and ScienceQA, which require strong multimodal reasoning capabilities. These results demonstrate that hierarchical feature integration not only enhances standalone vision tasks but also improves the LLM’s ability to process and reason over visual information.

\begin{table}[ht]
    \centering
    \caption{LLM-based evaluation results on vision-language tasks using SigLIP. Both SA and HIVE follow our three-stage pre-training method for the vision encoder. The final VLM training stage follows the procedure used in LLaVA. The best results are in bold.}
    \label{tab:llm-eval}
    \begin{tabular}{lcccc}
        \toprule
        Encoder & MME & GQA & OK-VQA & ScienceQA \\
        \midrule
        Base & 1296 & 57.74 & 48.78 & 62.34 \\
        SA & 1263 & 57.69 & 46.19 & 59.56 \\
        \rowcolor{gray!20} HIVE  
        & \cellcolor{gray!20} \textbf{1298} 
        & \cellcolor{gray!20} \textbf{58.05} 
        & \cellcolor{gray!20} \textbf{51.01} 
        & \cellcolor{gray!20} \textbf{63.12} \\
        \bottomrule
    \end{tabular}
\end{table}

\subsection{Gradient Map Visualization}
\label{subsec:gradient_analysis}

To understand how hierarchical cross-attention influences optimization, we analyze the gradient flow across vision encoder layers. This assessment is conducted through qualitative visualization.

\begin{figure}[ht]
    \centering
    \includegraphics[width=.85\linewidth]{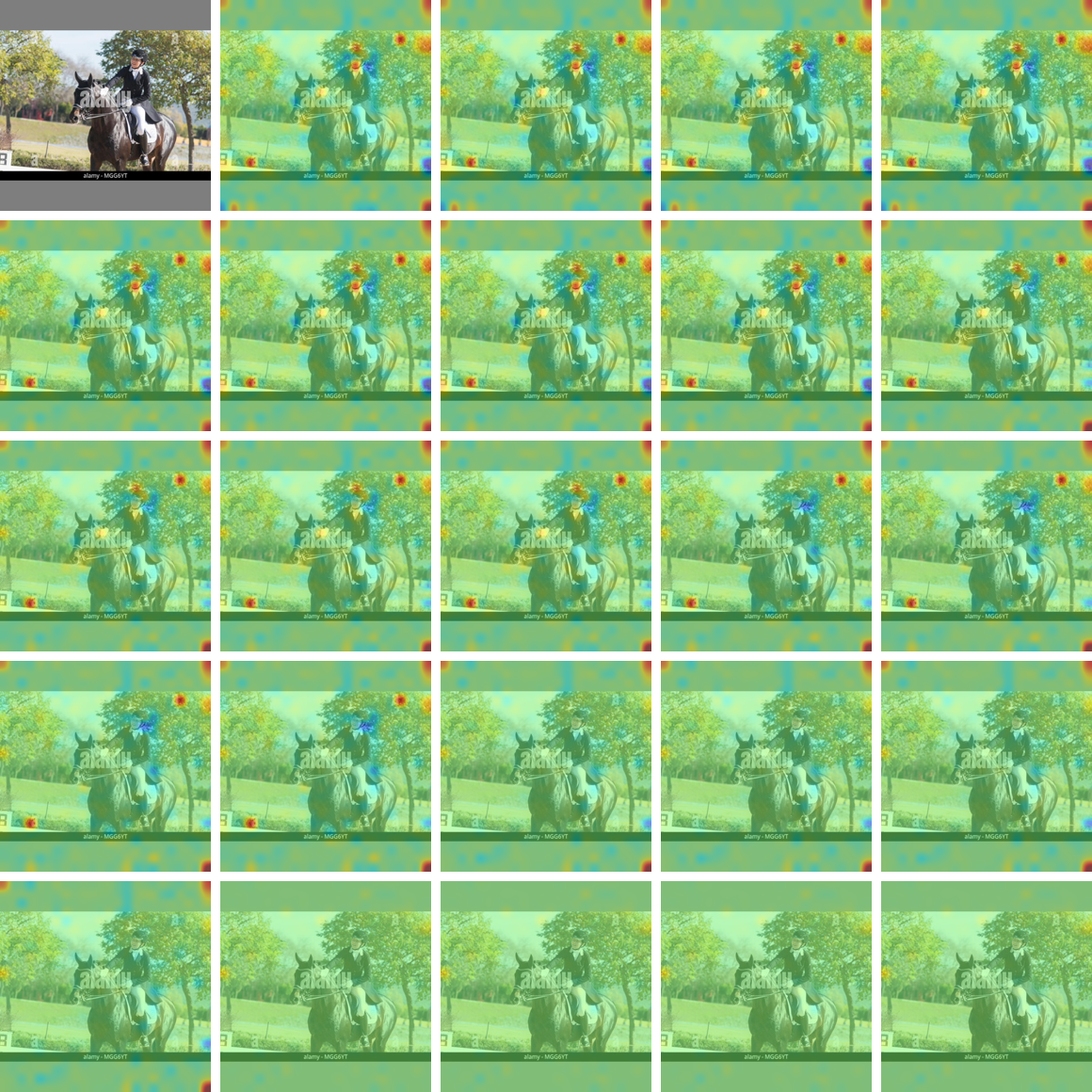}
    \caption{Gradient map visualization from the first to the final layer. The first image is the original image with the caption: ``a horse rider in full riding gear is riding on a horse - stock photo.'' The remaining images demonstrate more granular gradients in earlier layers due to cross-attention pre-training.}
    \label{fig:gradient_flow}
\end{figure}

We visualize the gradient distributions to examine how different regions of the model respond to backpropagation. Figure~\ref{fig:gradient_flow} presents the gradient maps from the first to the final vision encoder layer. The first image shows the original input image, captioned ``a horse rider in full riding gear is riding on a horse - stock photo.'' The subsequent images illustrate the gradient distribution across the encoder layers. 

Our results indicate that gradients in the earlier layers exhibit more granular patterns, which can be attributed to the improved feature refinement enabled by our cross-attention pre-training strategy. This granularity allows the model to better capture low-level features, enhancing the overall stability and effectiveness of hierarchical cross-attention. Qualitatively, we observe that these distinct, localized patterns in earlier layers contrast with the baseline. While not a definitive quantitative proof of stability, this visual evidence suggests that the cross-attention objective successfully forces gradients to flow back into early visual layers, a challenge often encountered in standard late-fusion models.

\subsection{Attention Map Visualization}
\label{subsec:attention_maps}

We analyze the attention maps of our model’s cross-attention layers to examine how hierarchical interactions influence visual feature selection. HIVE effectively leverages multi-level features, enhancing the LLM’s ability to align visual features with generated tokens.

\begin{figure}[ht]
    \centering
    \includegraphics[width=.85\linewidth]{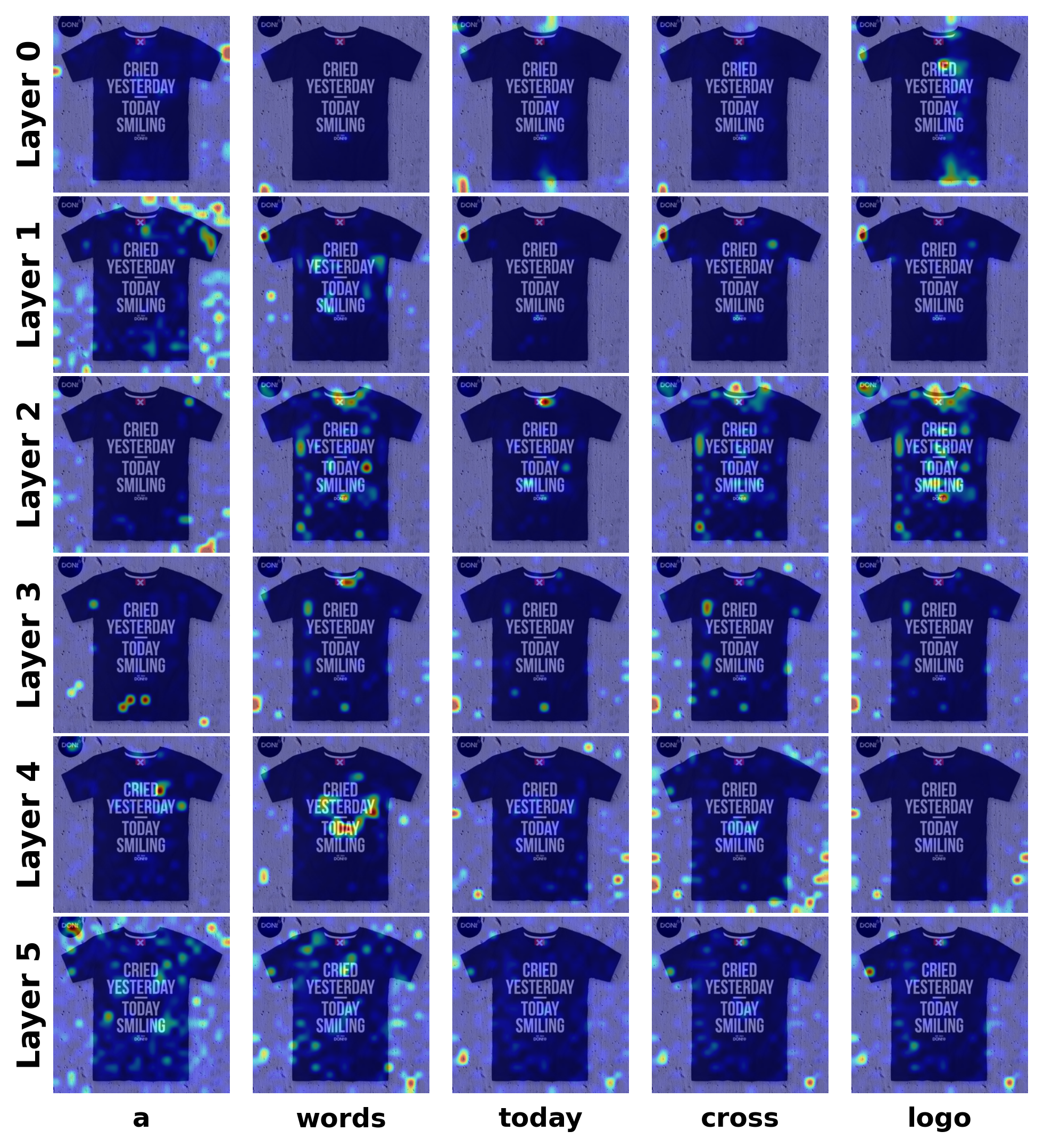}
    \caption{Attention map visualization illustrating hierarchical cross-attention behavior. The x-axis corresponds to sampled tokens, while the y-axis represents encoder layers from lower (top) to higher (bottom) layers. HIVE efficiently integrates multi-level features, with lower layers attending to specific details and higher layers producing broader, more abstract activations.}
    \label{fig:attention_maps}
\end{figure}

Our model demonstrates flexible attention behavior, dynamically adapting to both fine-grained details and broader semantic concepts. As shown in Figure~\ref{fig:attention_maps}, lower layers focus on specific regions, capturing textures, edges, and fine visual cues. In contrast, higher layers show more scattered attention patterns, reflecting abstract and global semantic concepts. This behavior arises from gradient flow aggregation, where lower layers accumulate refined visual cues through successive feature integration.

In Figure~\ref{fig:attention_maps}, we visualize five sampled tokens from the prompt: “a” (broad focus), “words,” “today,” “cross,” and “logo” (more specific targets). Early layers show localized, high-intensity activations, reflecting precise feature extraction. As depth increases, attention maps become more dispersed, aligning with broader visual contexts. This progressive refinement demonstrates HIVE’s advantage in combining detailed and abstract information, improving visual grounding and interpretability in vision-language tasks.

\subsection{Hierarchical Connection Strategy}
\label{subsec:ablation_hierarchy}

We evaluate the effect of our hierarchical cross-attention strategy through a brief study, without extensive ablations.

All experiments use a 25\% connection density, selected from preliminary observations to balance performance and efficiency. In CLIP, connections are uniformly distributed to leverage its class token for semantic aggregation. In SigLIP, they are applied to later layers, which better capture dispersed features due to the absence of a class token.

This strategy yields strong performance across both CLIP and SigLIP. While denser connections may offer slight gains, the 25\% setup effectively balances feature richness and computational cost.

\subsection{Runtime and Efficiency Analysis}
\label{subsec:runtime_analysis}

We compare the training efficiency of our hierarchical cross-attention framework against a full self-attention baseline, using the MobileLLM-350M model. While theoretical complexity is covered in Section~\ref{subsec:complexity_summary}, we provide empirical results on training time and memory use.

Hierarchical cross-attention offers notable efficiency gains, achieving a \textbf{3$\times$} speedup in per-epoch wall-clock training time and reducing peak GPU memory consumption by \textbf{55\%}, thereby improving scalability for large-scale training.

These results show that hierarchical cross-attention improves training efficiency without compromising vision-language performance. See Appendix~\ref{appendix:runtime_analysis} for detailed metrics.
\section{Conclusion}
\label{sec:conclusion}

We introduced HIVE, a hierarchical cross-attention framework that improves vision encoder pre-training for classification and vision-language tasks. By integrating multi-level features directly into the large language model, HIVE bypasses the limitations of standard late-fusion architectures, boosting performance while significantly lowering computational costs. Our empirical results demonstrate that HIVE consistently outperforms self-attention baselines, particularly in fine-grained visual recognition and complex multimodal reasoning. The substantial reduction in training time and memory overhead, achieved without sacrificing representational power, highlights HIVE’s potential as a highly efficient and scalable pre-training solution. Future work will explore extending this hierarchical integration to temporal modalities, such as video-language models, and investigating dynamic layer selection to further optimize cross-modal efficiency.

\clearpage

\section*{Acknowledgements}
This work was supported by the National Science and Technology Council of Taiwan (NSTC 114-2221-E-A49-157).

{
    \small
    \bibliographystyle{ieeenat_fullname}
    \bibliography{main}
}
\clearpage

\clearpage
\setcounter{page}{1}
\maketitlesupplementary

\begin{figure*}[t]
    \centering
    \includegraphics[width=\linewidth]{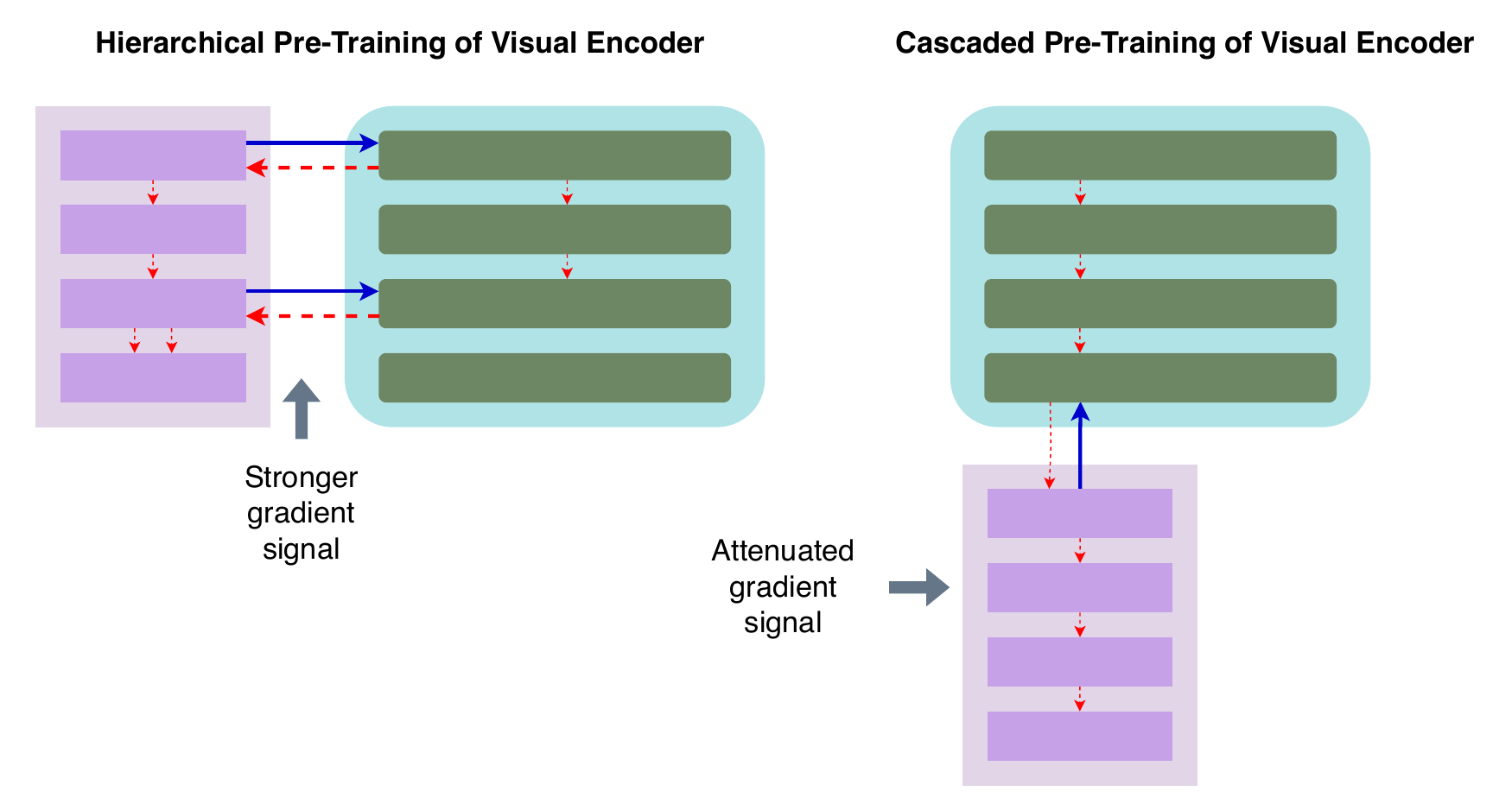}
    \caption{
        Comparison of hierarchical and cascaded pre-training approaches for vision encoders. 
        Hierarchical pre-training (left) establishes direct cross-attention across multiple layers, allowing for stronger gradient propagation and better feature integration. 
        Cascaded pre-training (right) restricts interactions to the final layer, leading to attenuated gradient signals and weaker hierarchical feature learning.
    }
    \label{fig:hive-overview}
\end{figure*}

\section{Hierarchical vs. Cascaded Pre-Training}
\label{appendix:hierarchical_vs_cascaded}

In this section, we compare hierarchical and cascaded pre-training approaches for vision encoders, highlighting their impact on gradient propagation and feature integration.

\paragraph{Hierarchical Pre-Training}  
Hierarchical pre-training establishes direct cross-attention between multiple layers of the vision encoder and the LLM. By enabling gradient flow across different levels of abstraction, this approach preserves fine-grained information while allowing deeper layers to refine high-level representations. As a result, hierarchical pre-training fosters better feature propagation and improves overall model convergence.

\paragraph{Cascaded Pre-Training}  
In contrast, cascaded pre-training follows a sequential learning process where only the final layer of the vision encoder interacts with the LLM. This approach reduces computational complexity but leads to attenuated gradient signals in earlier layers. Consequently, lower-layer features are less effectively incorporated into the model, potentially limiting performance in tasks requiring detailed visual understanding.

\paragraph{Key Observations}  
As illustrated in Figure~\ref{fig:hive-overview}, hierarchical pre-training offers superior gradient flow, leading to better optimization and improved feature learning. The cascaded approach, while computationally more efficient, may hinder the ability to leverage low- and mid-level vision features effectively.

Our experimental results (detailed in Section~\ref{sec:experiments}) demonstrate that hierarchical pre-training significantly enhances classification accuracy, particularly for vision-language tasks requiring fine-grained feature alignment. These findings suggest that hierarchical cross-attention is a more effective strategy for integrating vision encoders with LLMs.

\section{Hyperparameters}
\paragraph{Pre-Training} 
We outline the optimization hyperparameters and data augmentations used during HIVE pre-training in Table~\ref{tab:pretrain_hyperparams}. For tokenization, we adopt the tokenizer used by SigLIP~\cite{zhai2023sigmoid} and truncate any text longer than 77 tokens.

\begin{table*}[ht]
    \centering
    \caption{Pre-training hyperparameters for the three-stage pre-training procedure used in HIVE.}
    \label{tab:pretrain_hyperparams}
    \small
    \begin{tabular}{lccc}
        \toprule
        Stage & Stage 1 (Projector) & Stage 2 (Projector + LLM) & Stage 3 (Full Model) \\
        \midrule
        Optimizer & \multicolumn{3}{c}{Fully decoupled AdamW~\cite{loshchilov2017decoupled}} \\
        Optimizer Momentum & \(\beta_1 = 0.9, \beta_2 = 0.999\) & \(\beta_1 = 0.9, \beta_2 = 0.95\) & \(\beta_1 = 0.9, \beta_2 = 0.95\) \\
        Peak learning rate & \(1\times10^{-3}\) & \(2\times10^{-5}\) & \(2\times10^{-6}\) \\
        Minimum learning rate & \(1\times10^{-4}\) & \(2\times10^{-6}\) & \(0\) \\
        Weight decay & \(0\) & \(0\) & \(0\) \\
        Batch size & 256 & 256 & 1024 \\
        Epoch & 1 & 2 & 1 \\
        Gradient clipping & 1.0 & 10.0 & 10.0 \\
        Warmup iterations & 70 & 140 & 18 \\
        Total iterations & 2326 & 4652 & 581 \\
        Learning rate schedule & \multicolumn{3}{c}{Cosine decay~\cite{loshchilov2017decoupled}} \\
        \bottomrule
    \end{tabular}
\end{table*}

\paragraph{Classifier Fine-Tuning}
The optimization hyperparameters used during classifier fine-tuning are detailed in Table~\ref{tab:classifier_hyperparams}. For all experiments, we train a lightweight classifier on top of the frozen pre-trained vision encoder to evaluate the learned visual representations. This ensures a fair comparison across different pre-training approaches.

\begin{table*}[ht]
    \centering
    \caption{Classifier fine-tuning hyperparameters for HIVE and baselines.}
    \label{tab:classifier_hyperparams}
    \small
    \begin{tabular}{lc}
        \toprule
        Config & Setting \\
        \midrule
        Optimizer & AdamW~\cite{loshchilov2017decoupled} \\
        Optimizer Momentum & \(\beta_1 = 0.9, \beta_2 = 0.999\) \\
        Peak learning rate grid & \(2\times10^{-4}\) \\
        Minimum learning rate & \(0\) \\
        Weight decay & \(0.01\) \\
        Batch size & 512 \\
        Gradient clipping & 3.0 \\
        Warmup epochs & 1.5 \\
        Learning rate schedule & Cosine decay \\
        \midrule
        \multicolumn{2}{l}{\textit{Augmentations}} \\
        \midrule
        RandomResizedCrop & \\
        \quad Scale & [0.4, 1.0] \\
        \quad Ratio & [0.75, 1.33] \\
        \quad Interpolation & Bicubic \\
        RandomHorizontalFlip & \(p = 0.5\) \\
        ColorJitter & \\
        \quad Brightness & 0.2 \\
        \quad Contrast & 0.2 \\
        \quad Saturation & 0.2 \\
        \quad Hue & 0 \\
        \bottomrule
    \end{tabular}
\end{table*}

\paragraph{VLM Fine-Tuning for Vision-Language Tasks}
For vision-language model evaluation, we adopt a two-stage fine-tuning process based on the LLaVA~\cite{xu2024llava} framework:
\begin{itemize}
    \item \textbf{Stage 1: Connector Training.} We train the connector module to align the vision encoder’s output with the LLM token space while keeping both the vision encoder and LLM frozen.
    \item \textbf{Stage 2: LLM Fine-Tuning.} We freeze the vision encoder and train the LLM on downstream vision-language datasets to refine the model's reasoning capabilities.
\end{itemize}

The hyperparameters used in both stages are detailed in Table~\ref{tab:vlm_hyperparams}.

\begin{table*}[ht]
    \centering
    \caption{Hyperparameters for VLM fine-tuning on vision-language tasks.}
    \label{tab:vlm_hyperparams}
    \small
    \begin{tabular}{lcc}
        \toprule
        Config & Stage 1 (Connector Training) & Stage 2 (LLM Fine-Tuning) \\
        \midrule
        Optimizer & AdamW~\cite{loshchilov2017decoupled} & AdamW~\cite{loshchilov2017decoupled} \\
        Optimizer Momentum & \(\beta_1 = 0.9, \beta_2 = 0.999\) & \(\beta_1 = 0.9, \beta_2 = 0.999\) \\
        Peak learning rate & \(1\times10^{-3}\) & \(2\times10^{-5}\) \\
        Minimum learning rate & \(0\) & \(0\) \\
        Weight decay & \(0\) & \(0\) \\
        Batch size & 256 & 64 \\
        Gradient clipping & 1.0 & 1.0 \\
        Warmup iterations & 66 & 347 \\
        Total iterations & 2180 & 11540 \\
        Learning rate schedule & Cosine decay & Cosine decay \\
        \bottomrule
    \end{tabular}
\end{table*}

\section{Computational Efficiency}  
We evaluate the computational efficiency of hierarchical cross-attention compared to self-attention methods. Table~\ref{tab:efficiency_comparison} presents the measured training cost and memory overhead during model pretraining.  

Our method achieves improved efficiency by applying cross-attention to only \textbf{25\%} of the vision encoder layers, significantly reducing the number of attended tokens. Despite the reduced computational cost, our model consistently outperforms self-attention-based models across both classification and vision-language tasks (see Section~\ref{subsec:comparison}).

\begin{table}[ht]
    \centering
    \caption{Computational efficiency comparison between self-attention and HIVE. Values are reported relative to the self-attention baseline.}
    \label{tab:efficiency_comparison}
    \begin{tabular}{lc}
        \toprule
        \textbf{Method} & \textbf{Relative Efficiency} \\
        \midrule
        \textbf{Training Cost (TFLOPs)} & \\
        \ \ \ Self-Attention & 1.0× \\
        \ \ \ HIVE (Ours) & 0.14× \\
        \midrule
        \textbf{Memory Overhead} & \\
        \ \ \ Self-Attention & 1.0× \\
        \ \ \ HIVE (Ours) & 0.8× \\
        \bottomrule
    \end{tabular}
\end{table}

These results highlight the efficiency advantages of our method, which achieves improved performance despite lower computational cost during training.

\subsection{Computational Complexity Analysis}
\label{subsec:complexity}

In this section, we analyze the computational complexity of hierarchical cross-attention in comparison to full self-attention, particularly in the context of vision-language models (VLMs). Self-attention mechanisms are widely used in vision-language pretraining, but they introduce significant computational overhead when processing high-dimensional visual inputs. The proposed hierarchical cross-attention mechanism offers a more efficient alternative by selectively integrating multi-level vision features into the large language model (LLM), reducing redundant computations.

\paragraph{Self-Attention Complexity}  
In standard vision-language models, self-attention is applied across the full set of vision and text tokens. Given an input image \( \mathbf{I} \in \mathbb{R}^{H \times W \times C} \), the vision encoder tokenizes it into \( N_v \) visual tokens, where \( N_v = H W / P^2 \) and \( P \) is the patch size. The total number of tokens is given by:

\[
N = N_v + N_t
\]

where \( N_t \) is the number of text tokens. 

The complexity of full self-attention in the LLM is then:

\begin{equation}
    \mathcal{O} \left( L_l \frac{N^2 d}{2} + L_l N d^2 \right)
\end{equation}

where:
- \( L_l \) is the number of transformer layers in the LLM,
- \( N^2 / 2 \) arises from the causal masking in self-attention, which prevents future tokens from being attended to,
- \( d \) is the hidden dimension,
- The second term, \( \mathcal{O} (L_l N d^2) \), corresponds to the MLP complexity for processing all tokens in the transformer layers.

For high-resolution images, this quadratic term \( N^2 \) and extensive MLP computation become significant bottlenecks, making full self-attention costly for large-scale vision-language pretraining.

\paragraph{Hierarchical Cross-Attention Complexity}  
The proposed hierarchical cross-attention mechanism reduces computational cost by restricting interactions to a subset of vision encoder layers and selectively integrating multi-level features into the LLM. Instead of processing all \( N_v \) visual tokens at every layer, hierarchical cross-attention operates on a subset \( \mathcal{S} \) of informative layers from the vision encoder, each containing reduced feature representations.

Since visual tokens are not processed by the LLM's MLP layers, this design introduces substantial savings.

The complexity of hierarchical cross-attention is:

\begin{equation}
    \mathcal{O} \left( L_l L_s d^2 + L_l N_t d^2 \right)
\end{equation}

where:
- \( L_s \) is the number of selected vision encoder layers contributing to cross-attention,
- \( N_t \) is the number of text tokens that still pass through the LLM’s MLP layers.

Since \( L_s \ll N_v \), hierarchical cross-attention avoids the quadratic explosion of visual tokens seen in full self-attention and eliminates redundant MLP computations.

\paragraph{Comparison and Trade-offs}  
Hierarchical cross-attention significantly reduces computational overhead compared to full self-attention while maintaining strong performance across downstream visual tasks. The complexity comparison is summarized in Table~\ref{tab:complexity_comparison}.

\begin{table}[ht]
    \centering
    \caption{Computational complexity comparison between full self-attention and hierarchical cross-attention in vision-language models.}
    \label{tab:complexity_comparison}
    \begin{tabular}{lc}
        \toprule
        \textbf{Method} & \textbf{Complexity} \\
        \midrule
        Self-Attention (Causal) & \( \mathcal{O} \left( L_l \frac{N^2 d}{2} + L_l N d^2 \right) \) \\
        Cross-Attention (Causal) & \( \mathcal{O} (L_l L_s d^2 + L_l N_t d^2) \) \\
        \bottomrule
    \end{tabular}
\end{table}

Full self-attention provides maximum feature interactions but becomes impractical for high-resolution vision-language tasks due to its quadratic scaling with the number of visual tokens and extensive MLP computations. This results in high computational costs and memory overhead, limiting scalability. 

In contrast, hierarchical cross-attention selectively integrates multi-level visual features into the LLM, reducing redundant computations and significantly lowering computational requirements. By bypassing the LLM’s MLP layers for visual tokens, hierarchical cross-attention achieves substantial savings while maintaining strong performance in both visual and multimodal tasks. Experimental results demonstrate that cross-attention achieves superior performance across fine-grained and large-scale classification benchmarks, reinforcing its effectiveness in downstream visual tasks.

\subsection{Runtime and Efficiency Analysis}
\label{appendix:runtime_analysis}

We empirically evaluate the computational efficiency of our hierarchical cross-attention framework compared to a full self-attention baseline pretrained using the LLaVA 1B model. While theoretical complexity is discussed in Section~\ref{subsec:complexity}, here we measure actual runtime and memory usage to demonstrate practical scalability.

\paragraph{Experimental Setup}  
We measure the following metrics averaged over five runs:
\begin{itemize}
    \item \textit{Wall-clock training time}: Duration per training epoch.
    \item \textit{Memory consumption}: Peak GPU memory usage during training.
\end{itemize}

All evaluations use identical training parameters and batch sizes, ensuring fair comparisons.

\paragraph{Training Time Comparison}  
Table~\ref{tab:training_time} summarizes the average training epoch duration. Hierarchical cross-attention significantly reduces training time by limiting interactions to selected encoder layers.

\begin{table}[ht]
    \centering
    \caption{Average wall-clock training time per epoch. Hierarchical cross-attention provides notable speedups over self-attention.}
    \label{tab:training_time}
    \begin{tabular}{lcc}
        \toprule
        Model & Training Time (min/epoch) & Speedup \\
        \midrule
        Self-Attention & 240 & - \\
        Cross-Attention & 70 & 3.43$\times$ \\
        \bottomrule
    \end{tabular}
\end{table}

\paragraph{Memory Consumption}  
Peak GPU memory consumption during training is presented in Table~\ref{tab:memory_usage}. Our method considerably reduces memory requirements by decreasing token-level computations.

\begin{table}[ht]
    \centering
    \caption{Peak GPU memory usage during training. Hierarchical cross-attention reduces memory overhead substantially.}
    \label{tab:memory_usage}
    \begin{tabular}{lcc}
        \toprule
        Model & Peak Memory (GB) & Reduction \\
        \midrule
        Self-Attention & 54.2 & - \\
        Cross-Attention & 22.03 & 59.3\% \\
        \bottomrule
    \end{tabular}
\end{table}

\section{Gradient and Attention Map Visualizations}
\label{appendix:visualization}

In this section, we present qualitative visualizations of gradient flow and attention maps to illustrate the impact of hierarchical cross-attention on feature extraction and alignment.

\subsection{Gradient Flow Analysis}
\label{appendix:gradient_analysis}

\begin{figure}[ht]
    \centering
    \includegraphics[width=0.9\linewidth]{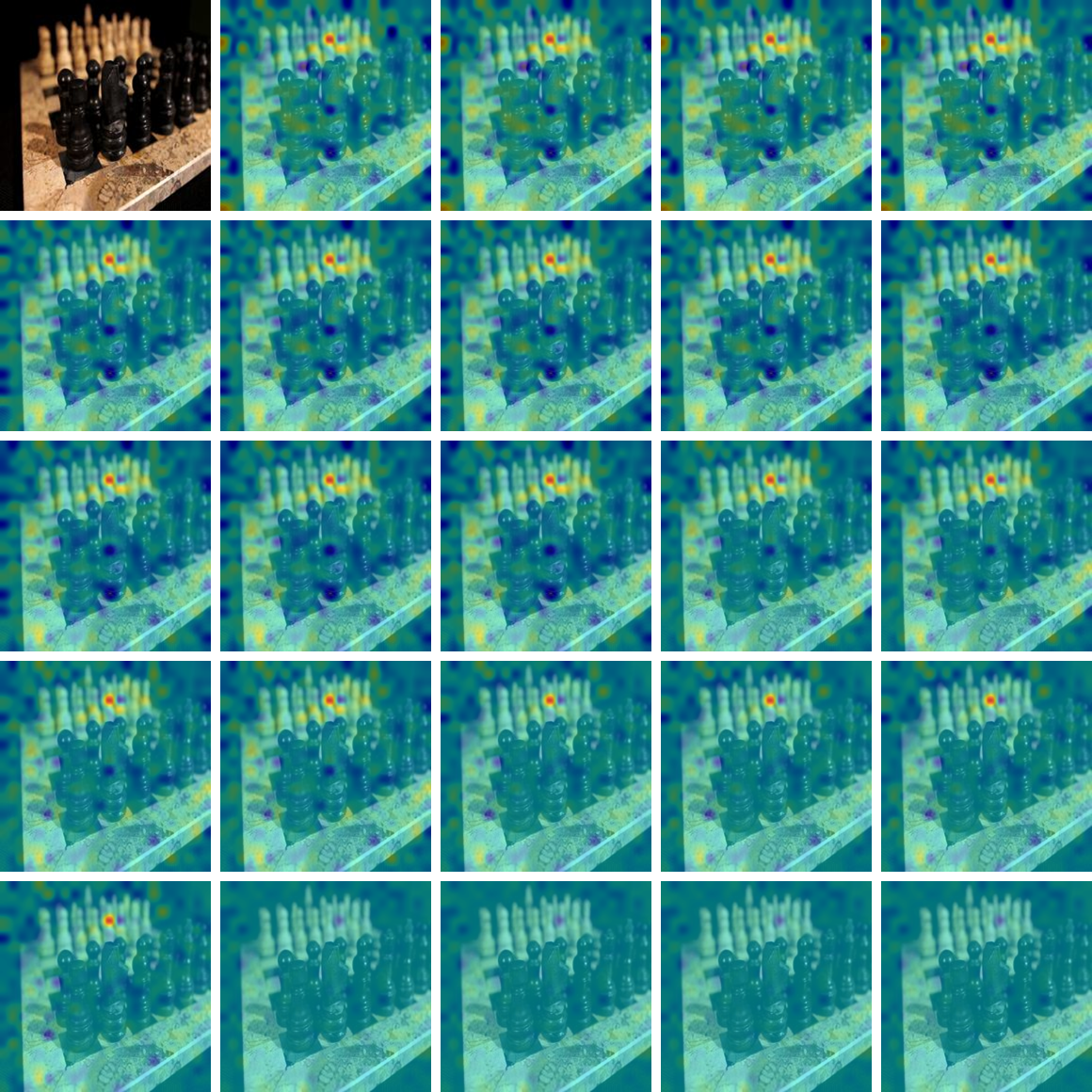}
    \caption{Gradient map visualization for Sample 1: "Investigators and journalists gather around the car of person after an attack on Wednesday." Gradients emphasize individuals and vehicles with sharp, localized activations in early layers.}
    \label{fig:gradient_flow_1}
\end{figure}

\begin{figure}[ht]
    \centering
    \includegraphics[width=0.9\linewidth]{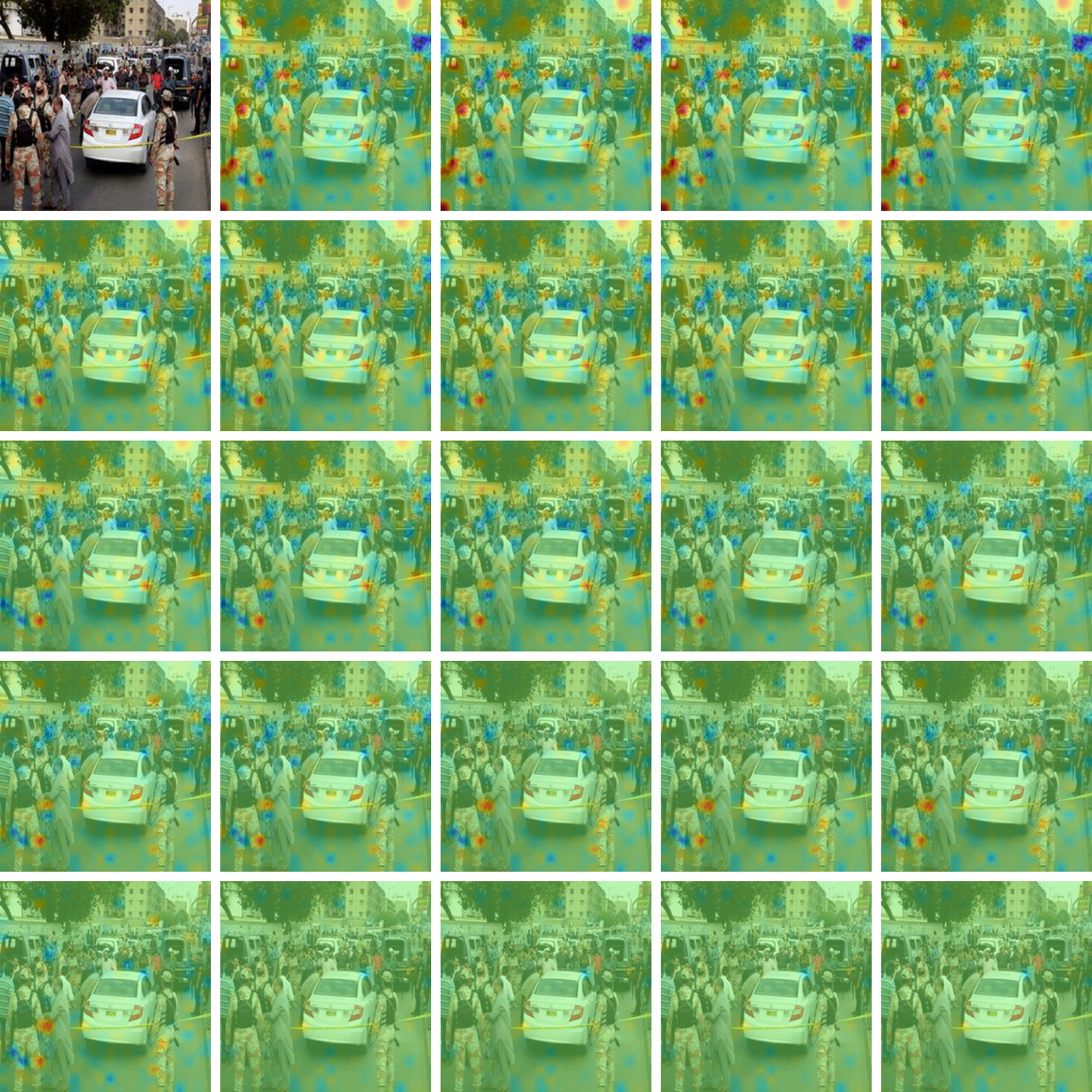}
    \caption{Gradient map visualization for Sample 2: "Colorful plastic and aluminum chairs leaning against tables at a cafe outdoor dining area." Early layers highlight fine-grained details such as chair edges, while deeper layers emphasize broader scene structure.}
    \label{fig:gradient_flow_2}
\end{figure}

\begin{figure}[ht]
    \centering
    \includegraphics[width=0.9\linewidth]{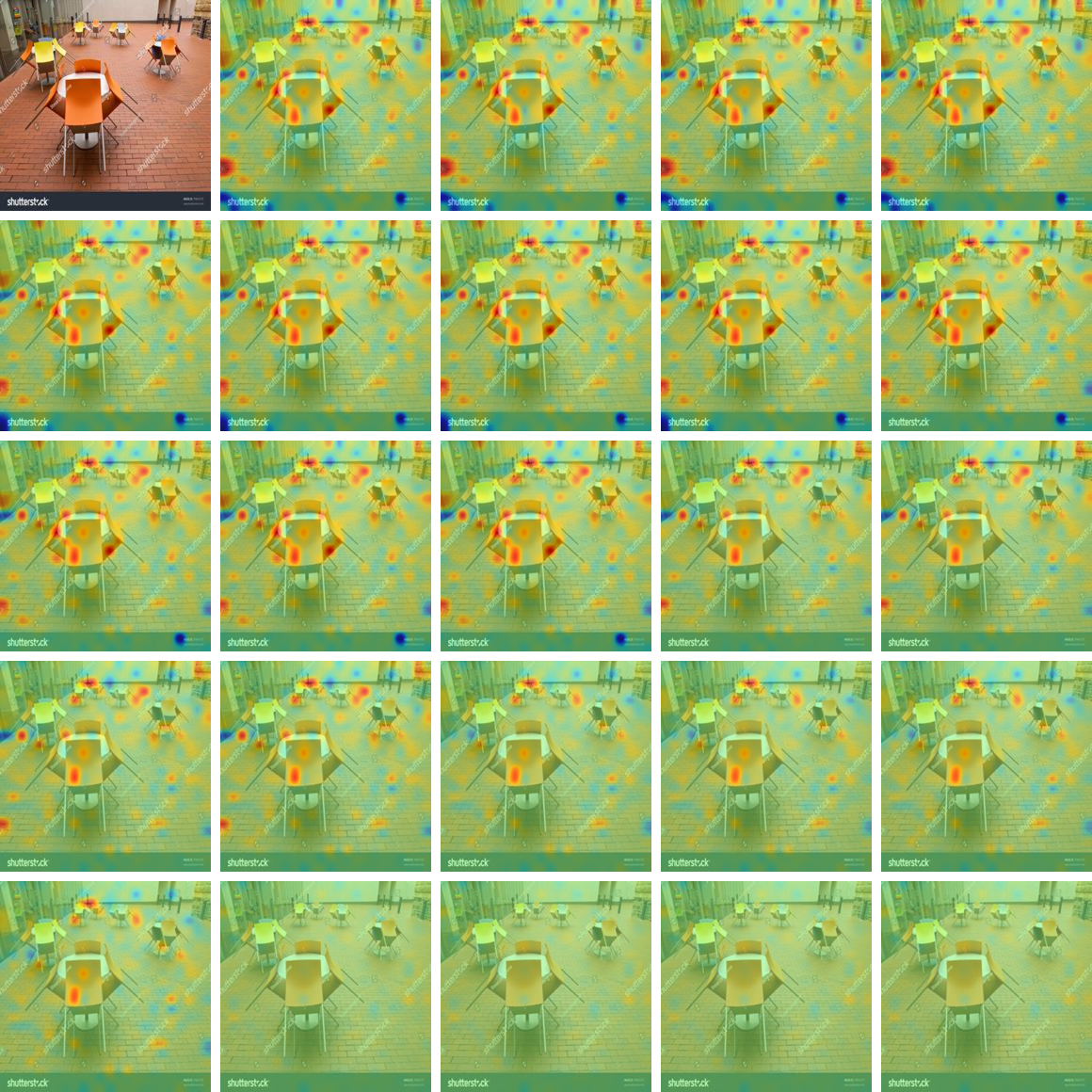}
    \caption{Gradient map visualization for Sample 3: "Person makes a move on defenders during the spring game." HIVE captures dynamic motion cues, focusing on the athlete and defenders.}
    \label{fig:gradient_flow_3}
\end{figure}

\begin{figure}[ht]
    \centering
    \includegraphics[width=0.9\linewidth]{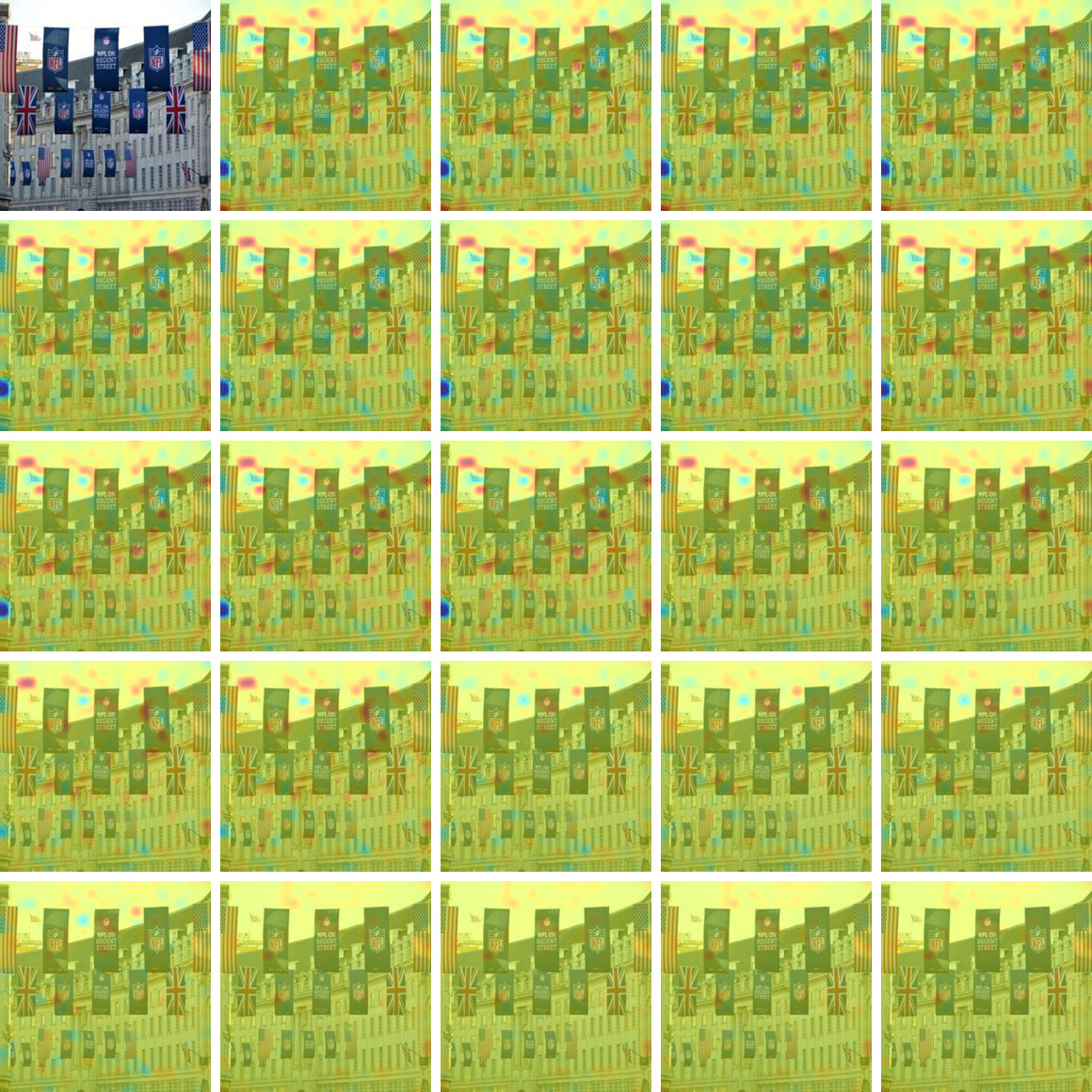}
    \caption{Gradient map visualization for Sample 4. HIVE maintains granular focus on key visual features, improving gradient flow.}
    \label{fig:gradient_flow_4}
\end{figure}

\begin{figure}[ht]
    \centering
    \includegraphics[width=0.9\linewidth]{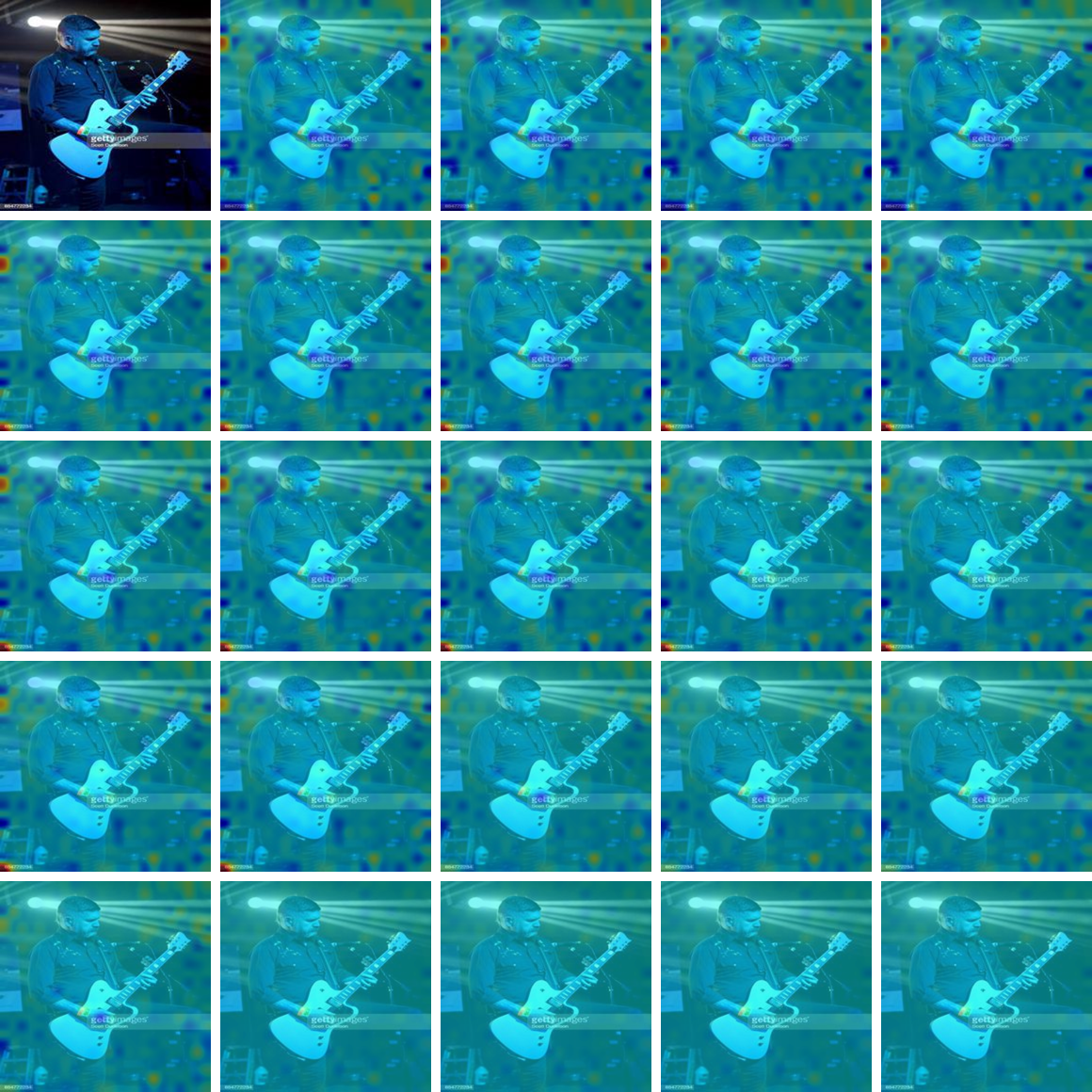}
    \caption{Gradient map visualization for Sample 5. Enhanced gradient stability enables sharper feature refinement in early layers.}
    \label{fig:gradient_flow_5}
\end{figure}

\begin{figure}[ht]
    \centering
    \includegraphics[width=0.9\linewidth]{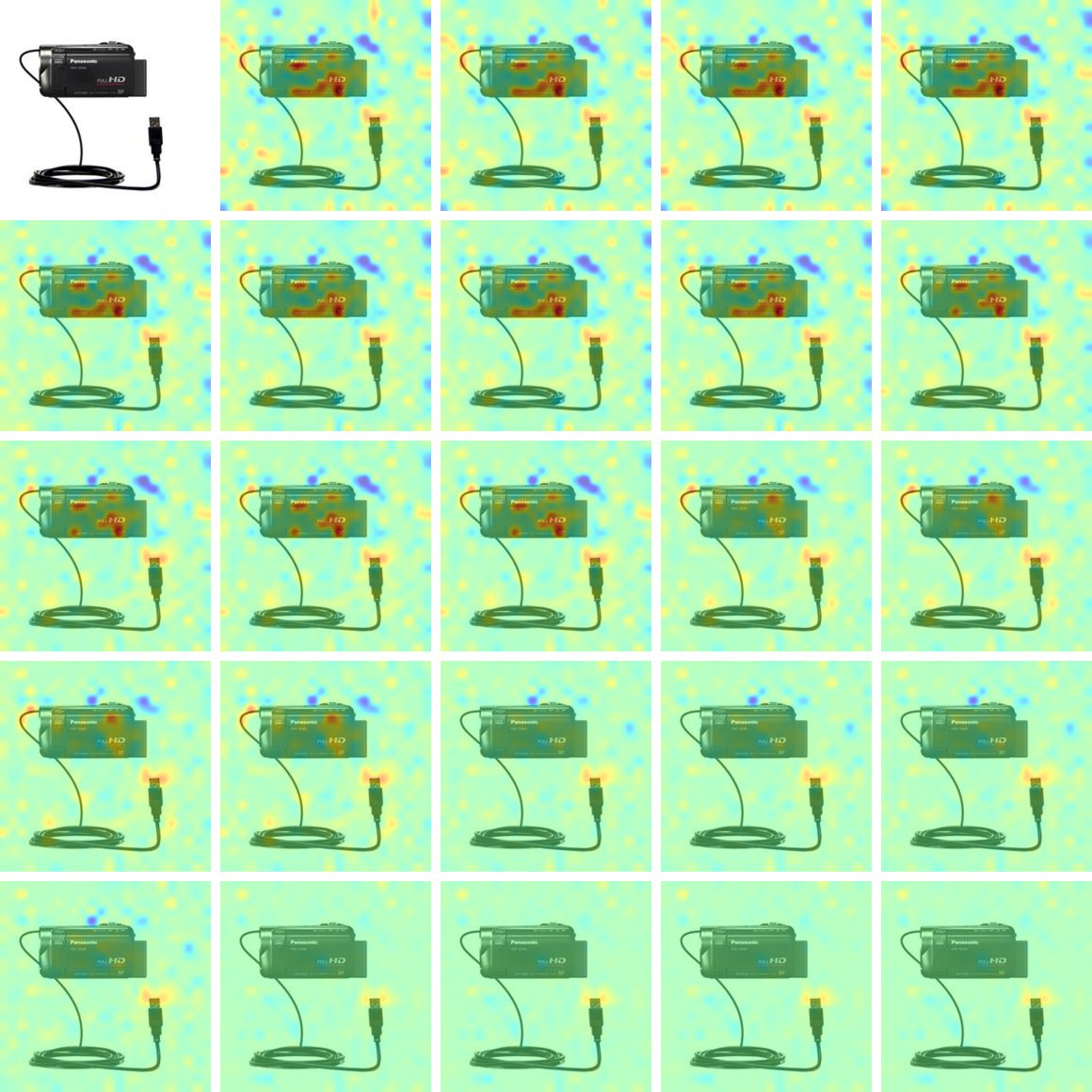}
    \caption{Gradient map visualization for Sample 6. HIVE consistently emphasizes meaningful visual elements across encoder layers.}
    \label{fig:gradient_flow_6}
\end{figure}

\begin{figure}[ht]
    \centering
    \includegraphics[width=0.9\linewidth]{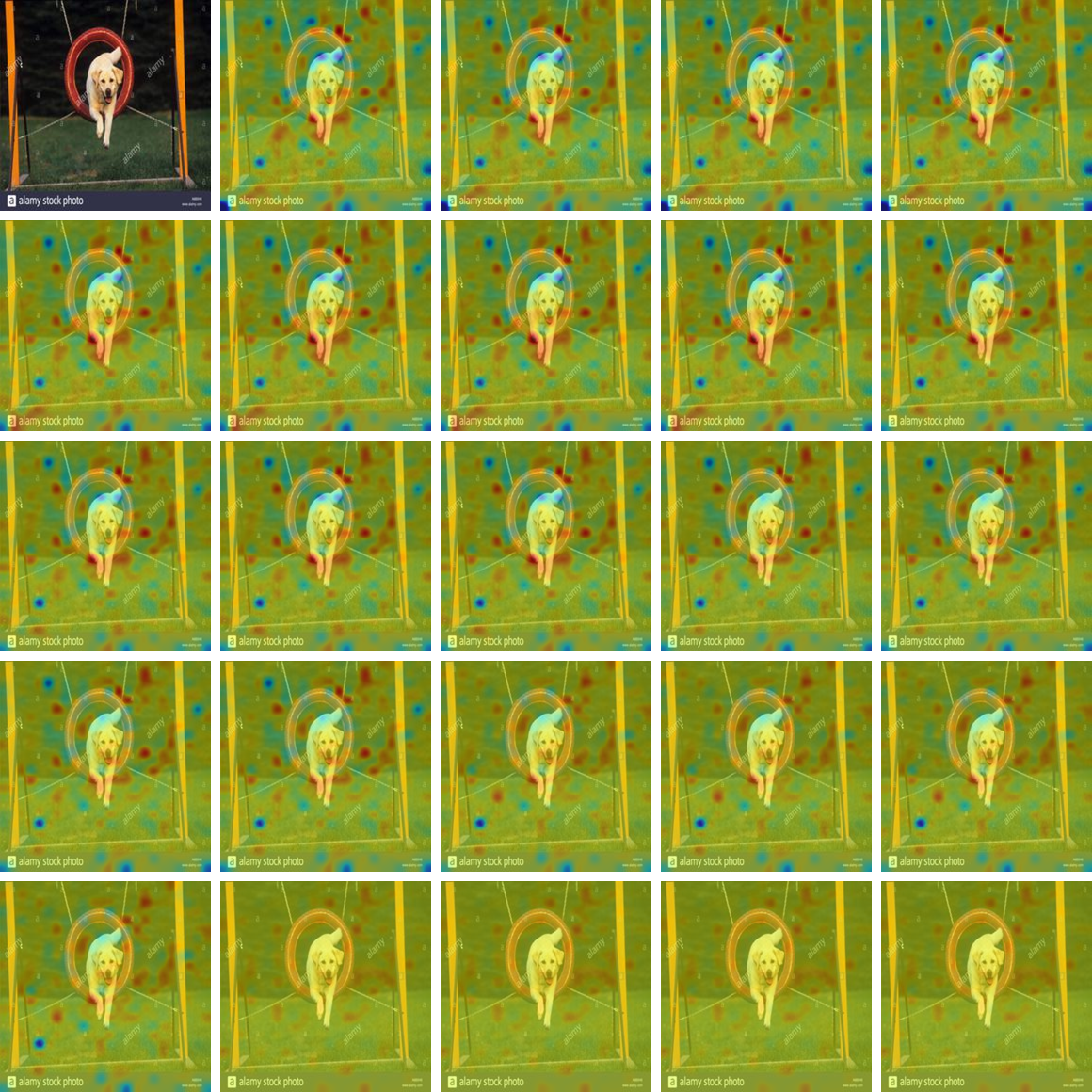}
    \caption{Gradient map visualization for Sample 7. Stable gradient propagation ensures effective visual feature learning across hierarchical layers.}
    \label{fig:gradient_flow_7}
\end{figure}

Figures~\ref{fig:gradient_flow_1} to \ref{fig:gradient_flow_7} present gradient map visualizations for various sample images. Each figure shows the original input (leftmost image), followed by gradient maps visualized across successive layers from the first to the final layer of the vision encoder.

\paragraph{Observations.}  
HIVE produces sharper and more structured gradient distributions compared to self-attention models. In earlier layers, gradients are highly granular, effectively capturing fine visual details such as textures, edges, and object boundaries. As the network deepens, the gradients become progressively broader, focusing on higher-level semantic regions. This behavior reflects the hierarchical nature of HIVE’s cross-attention mechanism, where low-level features are preserved while higher-level layers capture more abstract concepts.

The improved gradient flow is particularly evident in complex scenes, where key visual elements such as objects, text, and motion cues are effectively emphasized. This structured gradient propagation contributes to HIVE’s enhanced stability during training and improved visual representation learning.

\paragraph{Conclusion.}  
These visualizations highlight HIVE’s ability to promote stable gradient flow by efficiently distributing gradients across encoder layers. The observed improvements in feature refinement and gradient stability align with HIVE’s enhanced performance across vision-language and classification benchmarks.

\subsection{Attention Map Analysis}
\label{appendix:attention_maps}

\begin{figure*}[ht]
    \centering
    \includegraphics[width=\linewidth]{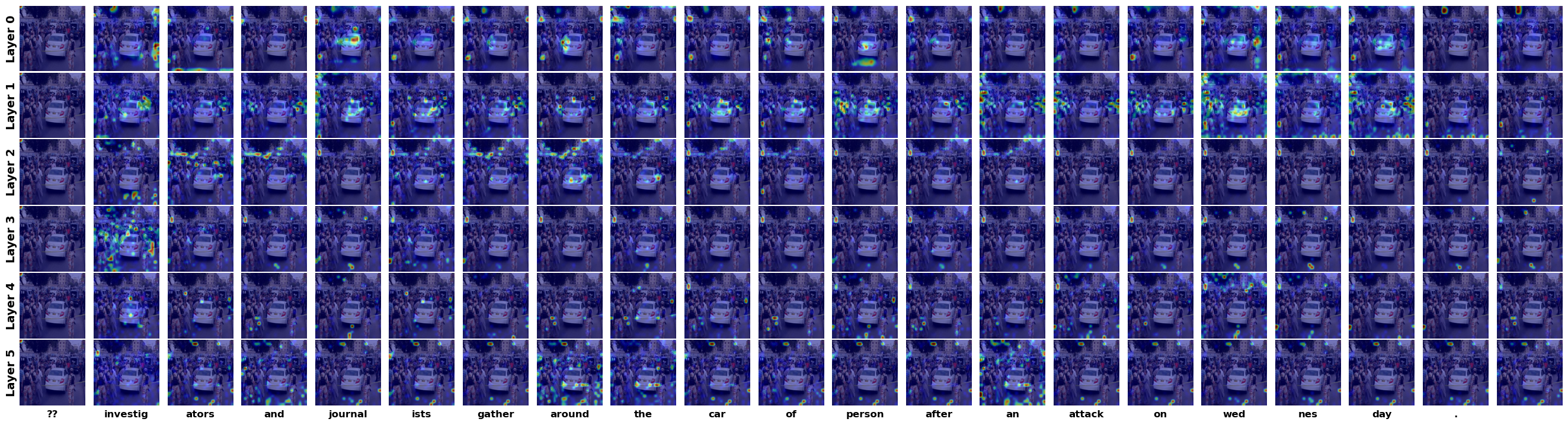}
    \caption{Attention map visualization for Sample 1: "Investigators and journalists gather around the car of person after an attack on Wednesday." HIVE emphasizes key elements such as the car and surrounding individuals, demonstrating improved semantic localization.}
    \label{fig:attention_maps_1}
\end{figure*}

\begin{figure*}[ht]
    \centering
    \includegraphics[width=\linewidth]{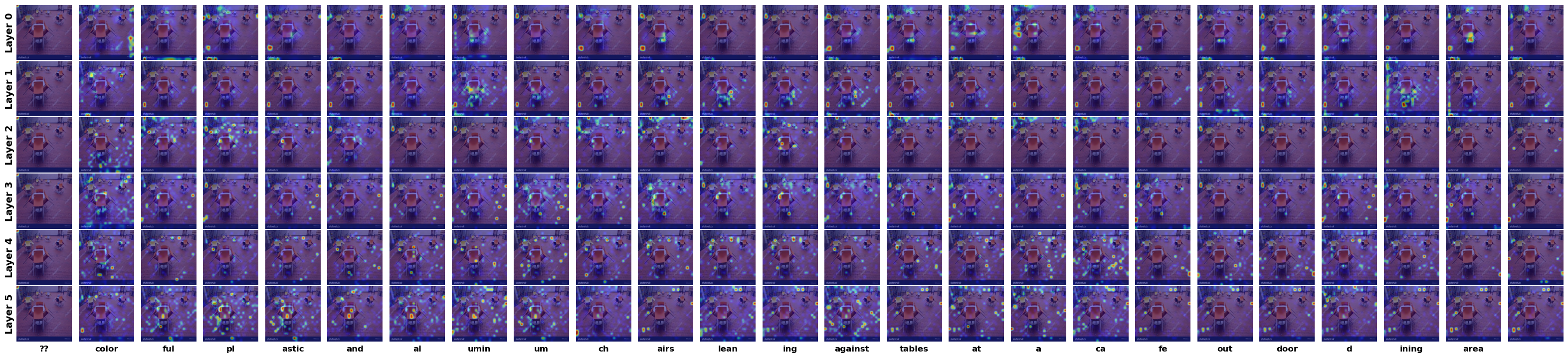}
    \caption{Attention map visualization for Sample 2: "Colorful plastic and aluminum chairs leaning against tables at a cafe outdoor dining area." HIVE highlights chairs and tables with sharper focus, enhancing object-level localization.}
    \label{fig:attention_maps_2}
\end{figure*}

\begin{figure*}[ht]
    \centering
    \includegraphics[width=\linewidth]{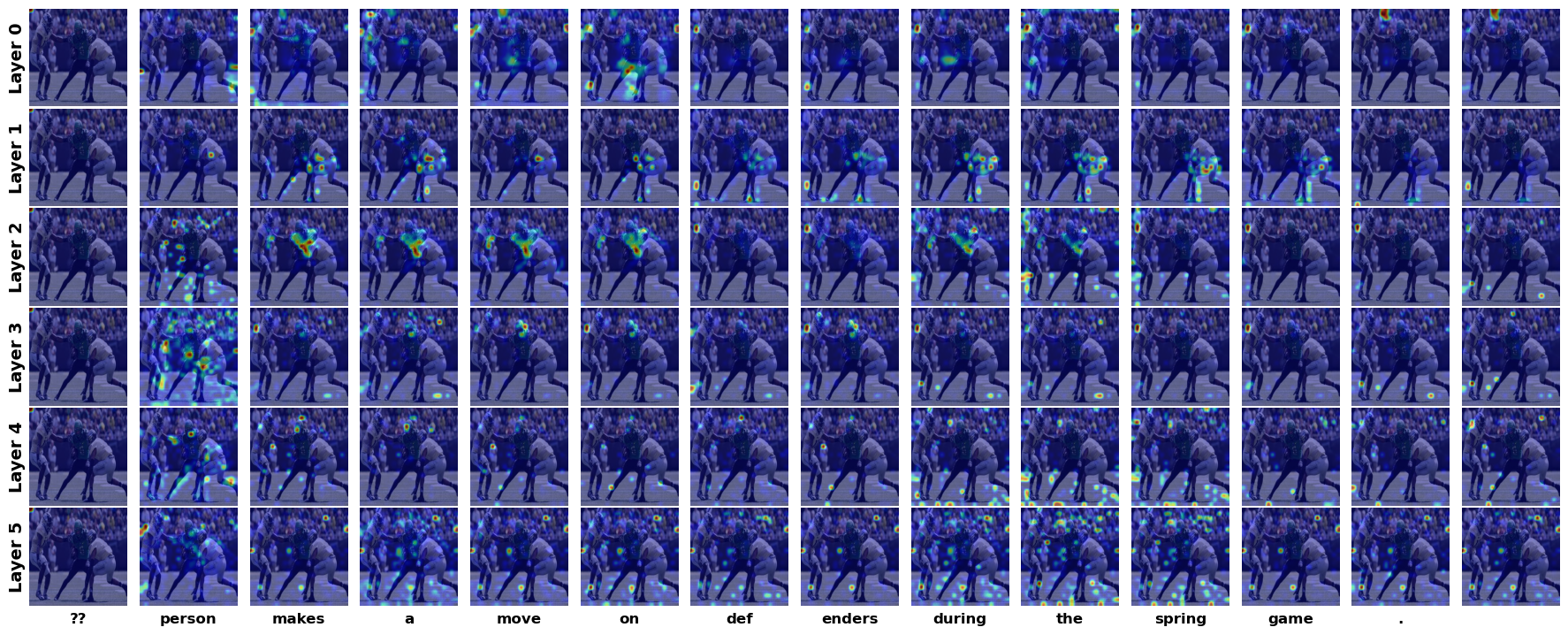}
    \caption{Attention map visualization for Sample 3: "Person makes a move on defenders during the spring game." HIVE effectively highlights the athlete’s movement and surrounding players, improving focus on dynamic elements.}
    \label{fig:attention_maps_3}
\end{figure*}

Figures~\ref{fig:attention_maps_1}, \ref{fig:attention_maps_2}, and \ref{fig:attention_maps_3} present visualizations of attention maps produced by HIVE's cross-attention layers. Compared to self-attention methods, HIVE’s cross-attention achieves sharper and more meaningful activations, improving token-to-region alignment.

In Figure~\ref{fig:attention_maps_1}, corresponding to the caption "Investigators and journalists gather around the car of person after an attack on Wednesday," HIVE effectively highlights key elements such as the car and surrounding individuals. The focused attention on these subjects illustrates HIVE’s ability to capture crucial semantic details in complex environments.

Figure~\ref{fig:attention_maps_2}, corresponding to the caption "Colorful plastic and aluminum chairs leaning against tables at a cafe outdoor dining area," shows HIVE's ability to isolate distinct objects, particularly the chairs and tables. The focused activations align with the scene's core visual features, emphasizing HIVE's improved object localization.

In Figure~\ref{fig:attention_maps_3}, corresponding to the caption "Person makes a move on defenders during the spring game," HIVE effectively emphasizes the athlete’s movement and surrounding players. This behavior highlights HIVE’s strength in capturing dynamic visual cues and distinguishing key elements in action-driven scenarios.

\paragraph{Conclusion.}  
These visualizations illustrate that HIVE’s cross-attention mechanism effectively integrates both low-level and high-level visual features. By dynamically attending to task-relevant regions, HIVE enhances visual grounding, improving performance across complex visual scenes in both static and dynamic environments.

\end{document}